\documentclass[sigconf,9pt]{acmart}
\makeatletter
\def\@ACM@checkaffil{
    \if@ACM@instpresent\else
    \ClassWarningNoLine{\@classname}{No institution present for an affiliation}%
    \fi
    \if@ACM@citypresent\else
    \ClassWarningNoLine{\@classname}{No city present for an affiliation}%
    \fi
    \if@ACM@countrypresent\else
        \ClassWarningNoLine{\@classname}{No country present for an affiliation}%
    \fi
}
\makeatother

\usepackage{amsmath, amssymb, amsfonts}
\usepackage{graphicx}
\usepackage{tabularx, booktabs, makecell, tabularray}
\usepackage{xspace}
\usepackage{xcolor}
\usepackage{enumitem}
\usepackage[caption=false, font=normalsize]{subfig}
\usepackage{algorithm}
\usepackage[noend]{algpseudocode}
\usepackage{lipsum}  
\usepackage{CJKutf8}
\usepackage{url}
\usepackage{natbib}
\usepackage{hyperref}

\AtBeginDocument{%
  \providecommand\BibTeX{{%
    \normalfont B\kern-0.5em{\scshape i\kern-0.25em b}\kern-0.8em\TeX}}}

\newcommand{\ie}{\emph{i.e.},\xspace}

\newcommand{\eg}{\emph{e.g.},\xspace}

\newcommand{\etal}{\emph{et al.}\xspace}

\newcommand\figref[1]{Fig.~\ref{#1}}

\newcommand\tabref[1]{Tab.~\ref{#1}}

\newcommand\secref[1]{Sec.~\ref{#1}}

\newcommand{\sysname}{{\sf AdaFlow}\xspace}
\newcommand{\sysnameposs}{{\sf AdaFlow's}\xspace}

\ifodd 1

\newcommand\lsc[1]{\textcolor{black}{#1}}

\else

\newcommand\lsc[1]{#1}

\fi

\begin{document}
\begin{CJK}{UTF8}{gbsn}

\title{AdaFlow: Opportunistic Inference on Asynchronous Mobile Data with Generalized Affinity Control}

\author{Fengmin Wu}
\email{fenny@mail.nwpu.edu.cn}
\affiliation{%
  \institution{Northwestern Polytechnical University}
  \country{}
}

\author{Sicong Liu}
\authornote{Corresponding author: scliu@nwpu.edu.cn}
\email{scliu@nwpu.edu.cn}
\affiliation{%
  \institution{Northwestern Polytechnical University}
  \country{}
}

\author{Kehao Zhu}
\email{zhukehao@mail.nwpu.edu.cn}
\affiliation{%
  \institution{Northwestern Polytechnical University}
  \country{}
}

\author{Xiaochen Li}
\email{lxcwuwang365@163.com}
\affiliation{%
  \institution{Northwestern Polytechnical University}
  \country{}
}

\author{Bin Guo}
\email{guob@nwpu.edu.cn}
\affiliation{%
  \institution{Northwestern Polytechnical University}
  \country{}
}

\author{Zhiwen Yu}
\email{zhiwenyu@nwpu.edu.cn}
\affiliation{%
  \institution{Northwestern Polytechnical University and Harbin Engineering University}
  \country{}
}

\author{Hongkai Wen}
\email{hongkai.wen@warwick.ac.uk}
\affiliation{%
  \institution{University of Warwick}
  \country{}
}

\author{Xiangrui Xu}
\email{xiangruixu@mail.nwpu.edu.cn}
\affiliation{%
  \institution{Northwestern Polytechnical University}
  \country{}
}

\author{Lehao Wang}
\email{lehaowang@mail.nwpu.edu.cn}
\affiliation{%
  \institution{Northwestern Polytechnical University}
  \country{}
}

\author{Xiangyu Liu}
\email{Liuxy01@mail.nwpu.edu.cn}
\affiliation{%
  \institution{Northwestern Polytechnical University}
  \country{}
}

\begin{abstract}
The rise of mobile devices equipped with numerous sensors, such as LiDAR and cameras, has spurred the adoption of multi-modal deep intelligence for distributed sensing tasks, such as smart cabins and driving assistance. 
However, the arrival times of mobile sensory data vary due to modality size and network dynamics, which can lead to delays (if waiting for slower data) or accuracy decline (if inference proceeds without waiting). 
Moreover, the diversity and dynamic nature of mobile systems exacerbate this challenge.
In response, we present a shift to \textit{opportunistic} inference for asynchronous distributed multi-modal data, enabling inference as soon as partial data arrives. 
While existing methods focus on optimizing modality consistency and complementarity, known as modal affinity, they lack a \textit{computational} approach to control this affinity in open-world mobile environments. 
\sysname pioneers the formulation of structured cross-modality affinity in mobile contexts using a hierarchical analysis-based normalized matrix. 
This approach accommodates the diversity and dynamics of modalities, generalizing across different types and numbers of inputs. 
Employing an affinity attention-based conditional GAN (ACGAN), \sysname facilitates flexible data imputation, adapting to various modalities and downstream tasks without retraining. 
Experiments show that \sysname significantly reduces inference latency by up to 79.9\% and enhances accuracy by up to 61.9\%, outperforming status quo approaches.
Also, this method can enhance LLM performance to preprocess asynchronous data.

\end{abstract}



\acmYear{2024}\copyrightyear{2024}
\acmConference[SenSys '24]{ACM Conference on Embedded Networked Sensor Systems}{November 4--7, 2024}{Hangzhou, China}
\acmBooktitle{ACM Conference on Embedded Networked Sensor Systems (SenSys '24), November 4--7, 2024, Hangzhou, China}
\acmDOI{10.1145/3666025.3699361}
\acmISBN{979-8-4007-0697-4/24/11}

\maketitle

\section{Introduction}
\label{sec:intro}

The broad integration of low-power, rich-data sensors like cameras, LiDAR, acoustic sensors, hyperspectral cameras, and radio frequency detectors has greatly advanced the sensing potential of mobile devices~\citep{bib:mml:b,bib:er:l}.
Leveraging multiple sensors for multi-modal inference provides robust perception outcomes compared to single-sensor systems across various domains, such as smart cockpits (\eg Google automotive service~\citep{bib:gas:g}), 
driving assistance (\eg Tesla Autopilot~\citep{bib:tesla:t}),
3D scene understanding (\eg Microsoft Azure Kinect DK~\citep{bib:kinect:m}), health monitoring (\eg Apple HealthKit~\citep{bib:apple:a}), and AR/VR (\eg Meta Oculus Rift~\citep{bib:face:a}).
The advantages of \textit{distributed} multi-modal systems are particularly evident 
in challenging environments or objects. 
For instance, observing occluded objects, penetrating through walls, operating in low-light conditions, and navigating around corners. 
The diversity of sensor data from various fields of view and sensitivities supports independent cross-validation from multiple vantage points and enables detailed extraction of complex patterns (\eg location, direction, and materials). 

A core issue in distributed multi-modal systems is how to process the vast, continuously generated data streams from advanced sensors. Many studies focus on technologies and tools for handling multi-modal data, such as adopting AdaptSegNet for fusing complex signals~\citep{bib:ssm:v}, or further explored partitioning models across mobile devices and the cloud to enhance processing speed~\citep{bib:neu:k}.

However, the significant \textit{gap} exists in open-world mobile deployments, characterized by input \textit{asynchrony} and \textit{heterogeneity}:

\noindent$\bullet$ \textbf{Asynchronism}. Different input modalities (\eg video \textit{vs.} LiDAR) and dynamic network conditions lead to asynchronous data arrival at the fusion server. 
    Also, low-power mobile sensors often lack clock synchronization.
    This may lead to delays (when waiting for slow data), or accuracy decline (when fusion proceeds without waiting). 
    For example, in real-world autonomous driving scenarios with 6 cameras and 1 LiDAR, a 40ms asynchronous delay occurs in LiDAR data over a 100Mbps bandwidth, given $1:4$ data size between LiDAR and camera data.
    This delay increases latency from 1s to 4.3s for 81 frames (in a $4s$ window) when waiting for slow data, whereas proceeding without waiting degrades accuracy by 49.7\% (we defer more details in \secref{subsec:problem}). 
    However, \lsc{mobile systems must minimize latency (\eg $\leq 26ms$ per frame) to ensure safety and operational efficiency~\citep{bib:dl:m}.} 

\noindent$\bullet$ \textbf{Heterogeneity}. Variations in the field of view, sensitivity, noise (\eg lighting), and feature distribution of distributed multi-modal data affect the \textit{consistency} and \textit{complementarity} 
    of fusion results. 
    Improper management of these factors when deciding whether to discard or impute slow modalities can impact inference accuracy and latency.
    For example, using real-world autonomous driving dataset nuScenes~\citep{bib:nuscenses:ca}, discarding 50\% of the slow data which exhibits approximately 33\% arrival delays, results in an 28.1\% decrease in accuracy. 
    Using KNN~\citep{bib:knn:c}, a typical asynchronous data imputation method, reduces the issue but leads to a 9.7\% accuracy drop and a 328\% increase in latency (see \secref{subsec:observation}).

As a separate note, cross-modal \textit{consistency} and \textit{complementarity}, collectively referred to as modality \textit{\textbf{affinity}}, explore inference robustness and accuracy respectively by leveraging \textit{modality-shared} and \textit{modality-specific} information, respectively~\citep{bib:cosmo:oyxm}.
Modality affinity changes dynamically with input asynchronism and heterogeneity, variable across tasks, even with the same inputs (see \secref{subsec:observation}).
Given the above characteristics, existing synchronous (blocking)~\citep{bib:mml:b, bib:multimodal:hong, bib:multimodal:liu} and asynchronous (non-blocking)~\citep{bib:lls:L} techniques for distributed multi-modal inference face following challenges:

\noindent$\bullet$ \textbf{\textit{Challenge $\#$1}}: 
Most prior multi-modal methods optimize data \textit{consistency} and \textit{complementarity} to balance accuracy and latency. 
However, none offer a \textit{quantitative} structured affinity analysis among modalities for precise assessment and control.
Also, most methods explore cross-modal affinity only \textit{qualitatively} and under \textit{static} or \textit{ideal} noise-free conditions~\citep{bib:taskonomy:AZ, bib:prov:z}.

\noindent$\bullet$ \textbf{\textit{Challenge $\#$2}}: 
Scheduling asynchronous and heterogeneous multi-modal data for low-latency inference is non-trivial, involving NP-hard multi-choice, multi-strategy optimization problems~\citep{bib:taskonomy:AZ, bib:adam:k, bib:rmsprop:t}.
Moreover, a one-fit-all imputation module is essential yet challenging to handle \textit{dynamic} and \textit{diverse} input modalities, various subsequent tasks, and fluctuating asynchronous delays.

To address these challenges, we propose a shift to \textit{opportunistic} inference, where the server performs clever inference as soon as partial asynchronous data are available.
\sysname harnesses modality affinity to ensure robust data fusion at runtime. 
It is grounded in a fully computational method that models and optimizes inference based on modality affinity.

\noindent$\bullet$ \textit{\textbf{First}}, drawing inspiration from~\citep{bib:taskonomy:AZ}, which explores \textit{task} affinity using \textit{unimodal} (visual) data during \textit{learning}, we extend this concept to the \textit{modality} level during \textit{inference}, enabling dynamically adaptive assessments of diverse modalities. 
Furthermore, \sysname employs the analytic hierarchy process (AHP) to normalize matrix values, accommodating heterogeneity and asynchrony.

\noindent$\bullet$ \textit{\textbf{Second}}, to enable low-latency and high-accuracy non-blocking inference that adapts to varying input modalities and types without retraining at runtime, we introduce a one-fit-all attention-based conditional generative network (ACGAN). 
This network adaptively embeds input features and imputes data, allowing for precise and timely control of modality affinity in opportunistic inference, paving the way for responsive mobile systems.

We evaluate the performance of \sysname on real-world 3D object recognition tasks and scenarios with diverse data heterogeneity and asynchrony. 
Results show a reduction of up to 79.9\% in inference time with an improvement of up to 61.9\% in accuracy.
Especially, it displays the affinity matrix enables opportunistic inference based on different downstream tasks and even scenarios.
This method can also be a pre-module for LLM, enhancing performance on asynchronous data.
Our main contributions are as follows.
\begin{itemize}
    \item As far as we know, this is the first work to integrate quantitative modality affinity control into distributed multi-modal inference, addressing \textit{system asynchrony}. 
    It eliminates waiting times for slow data and minimizes accuracy loss when excluding such data.
    \item We introduce \sysname, a system for opportunistic inference that adapts to asynchronous and heterogeneous data streams. 
    It harnesses structured affinity control to enable adaptive fusion of dynamic data. 
    Also, \sysname proposes an ACGAN to achieve precise and timely data imputation.
    \item Experiments show that \sysname outperforms existing a-/synchronous methods~\citep{bib:patch:w,bib:mdl:n,bib:pcn:Y,bib:knn:c} in trading off between inference accuracy and latency across various real-world mobile tasks, inputs, and scenarios. 
\end{itemize}

\section{Overview}

\subsection{Problem Analysis}
\label{subsec:problem}



\begin{figure}[t]
    \centering 
    \subfloat[]{\label{fig:exp_alpha}
    \includegraphics[height=0.22\linewidth]{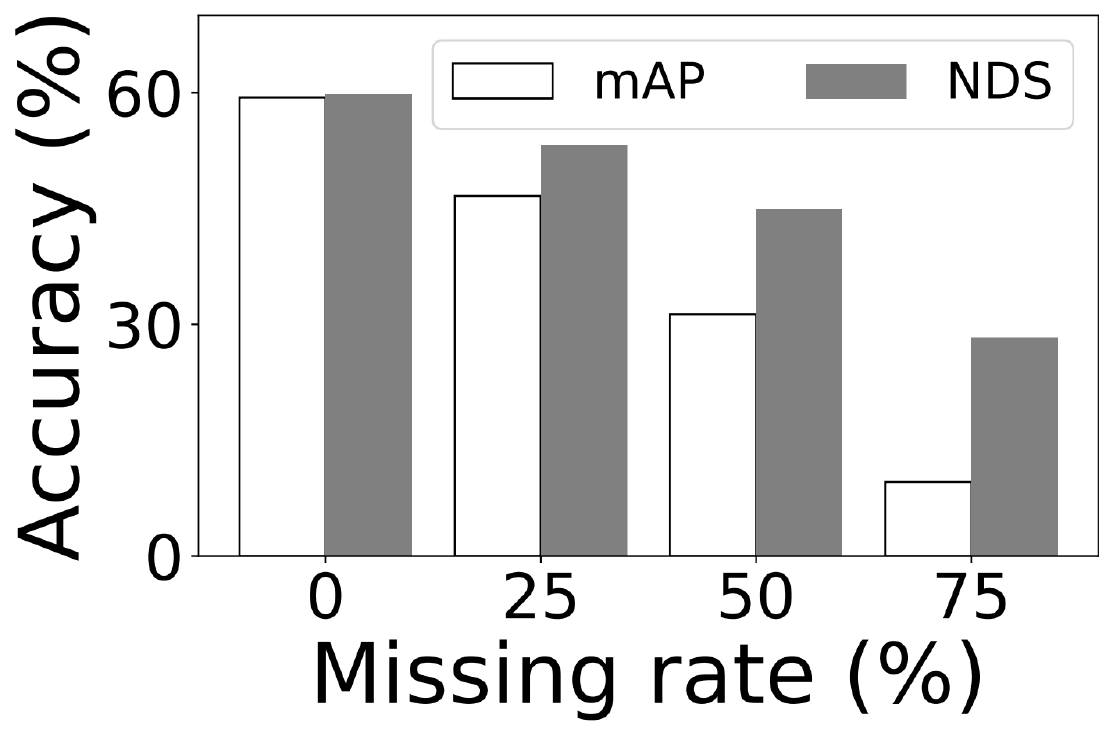}}
    \hspace{5mm}
     \subfloat[]{\label{fig:exp_lr}
    \includegraphics[height=0.22\linewidth]{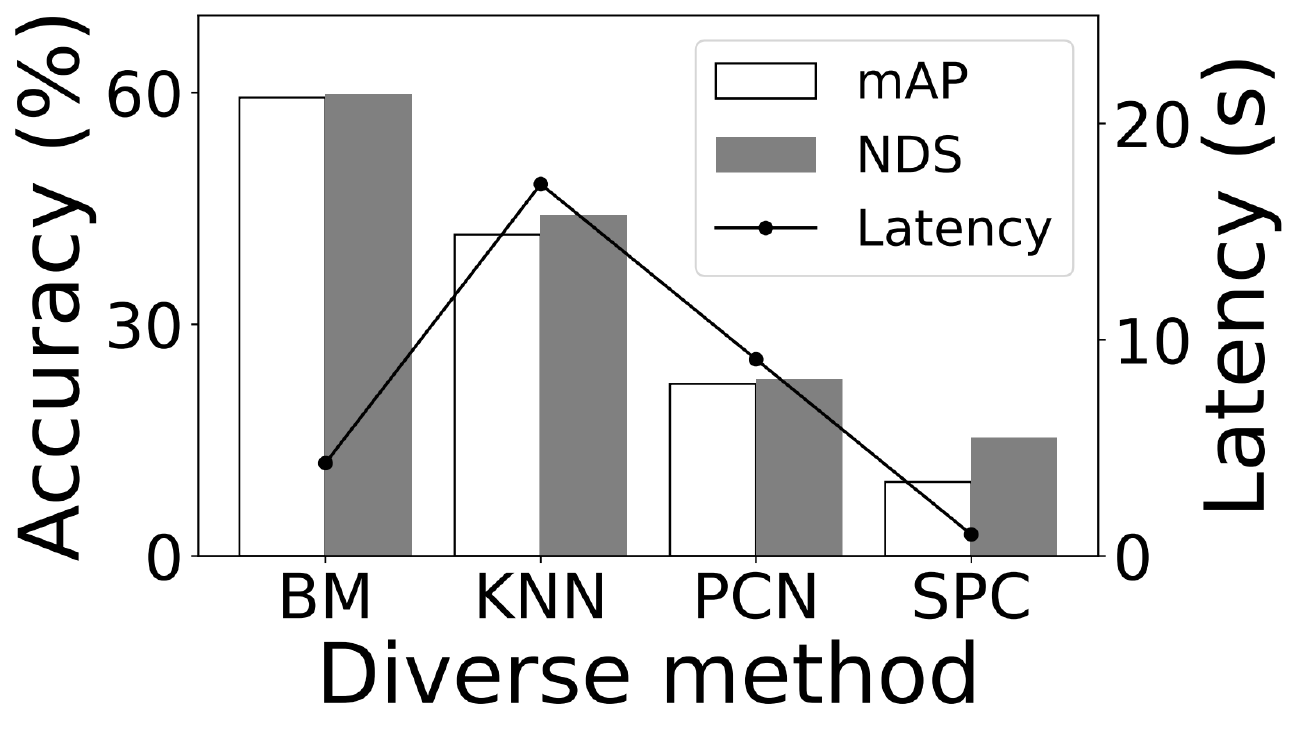}}
    \caption{(a) Inference accuracy, (b) accuracy \& latency of existing syn-/asynchronous methods, under varying missing rates of slow modality (\ie 3D point cloud).}
\label{fig-motivation}
\end{figure}

To analyze and observe the motivation, we conducted the following experimental tests.
We test a distributed multi-modal 3D object recognition system on nuScenses dataset~\cite{bib:nuscenses:ca}.
nuScenses, collected from real-world autonomous driving scenarios, includes 6 camera and 1 LiDAR views. 
Given the 1:4 data size ratio between LiDAR and camera data, a 40ms asynchronous delay occurs in LiDAR data on a 100Mbps bandwidth.
\textit{First}, as illustrated in \figref{fig-motivation}a, we simulate data \textit{asynchronism} using diverse missing rates of slow modality (\ie 3D LiDAR point cloud), \ie 0\%, 25\%, 50\%, and 75\%. 
As the missing rate of the slow modality increases, while the fast modality (\ie RGB images) remains fully available, if not waiting for slow data, the inference accuracy progressively declines (\eg 49.8\%).
If waiting for slow data, the latency increases from 1s to 4.3s for 81 frames.
\textit{Second}, as shown in \figref{fig-motivation}b, we test various imputation methods for \textit{asynchronous fusion} such as SPC (Sparse Point Cloud)~\cite{bib:mdl:n}, KNN (K-Nearest Neighbors) \cite{bib:knn:c}, and PCN (Point Completion Net) \cite{bib:pcn:Y}, along with the \textit{synchronous} fusion method BM (Blocking Mechanism)~\cite{bib:patch:w}, across missing rates of 0\%, 25\%, 50\%, and 75\% in the slow data (\ie 3D point cloud) when the fast data (\ie RGB image) is available.
We find that while the blocking method offers the highest accuracy, its latency, nearly equal to the sampling interval of the test data (4s), is too high. Conversely, the sparse point cloud method is the fastest but yields an unacceptable accuracy drop (\ie 4.29\%).

\subsection{Observations and Opportunities}
\label{subsec:observation}
In mobile applications, a distributed multi-modal data fusion system aims to optimize the tradeoff between inference accuracy and latency. 
A significant challenge with existing a-/synchronous methods (referenced in \secref{sec:related}) is their failure to assess and manage quantitatively \textit{modality affinity}, encompassing both modality consistency and complementarity, leading to delays or accuracy drop. 
Despite its importance, modality affinity remains a \textit{black box}, exacerbated by the unpredictable asynchrony and heterogeneity in mobile contexts. 
To address this, we conduct specific tests and make the following observations.

\begin{figure}[t]
    \centering 
    \subfloat[]{\label{full}
    \includegraphics[height=0.4\linewidth]{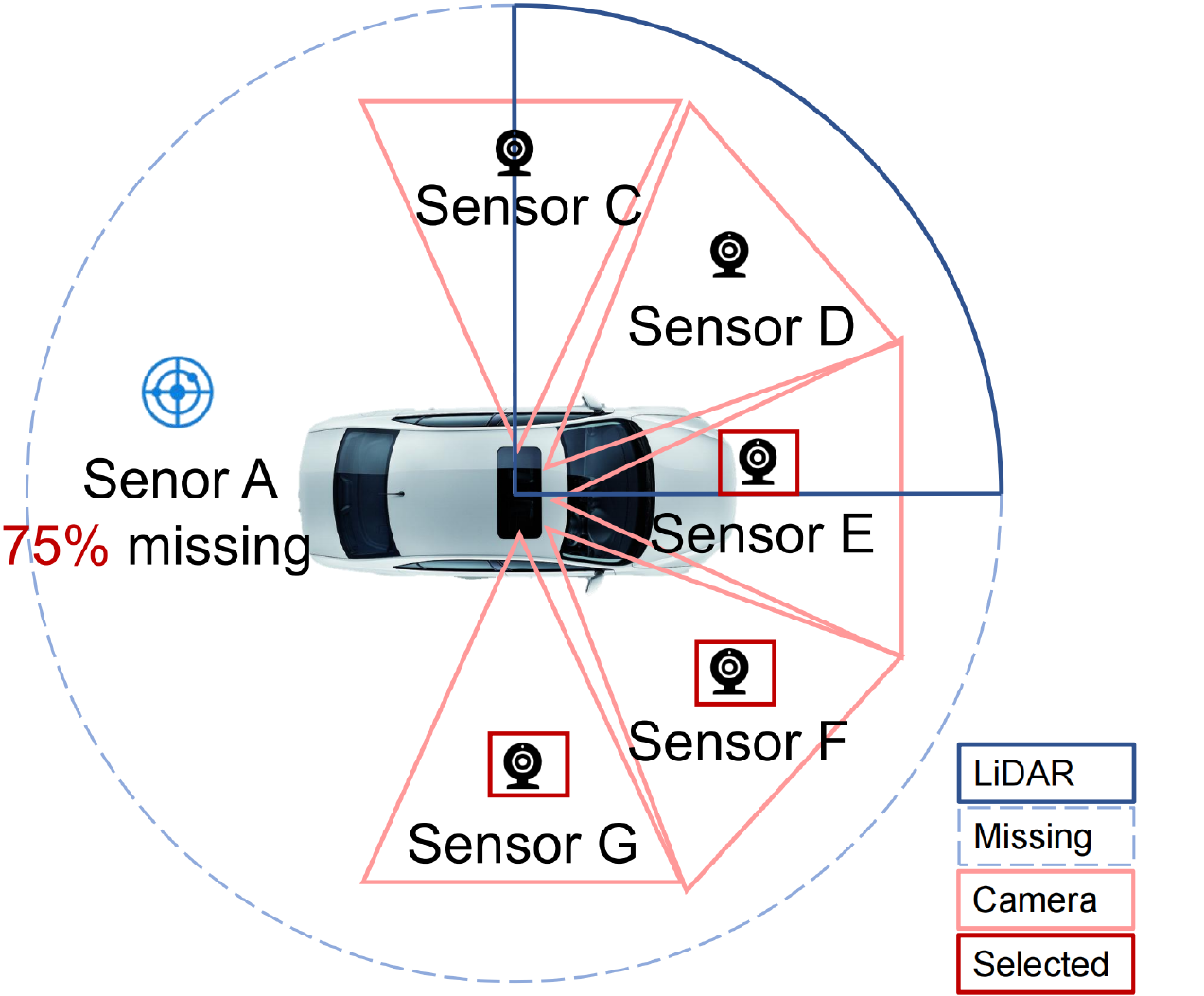}}
     \subfloat[]{\label{partial}
    \includegraphics[height=0.4\linewidth]{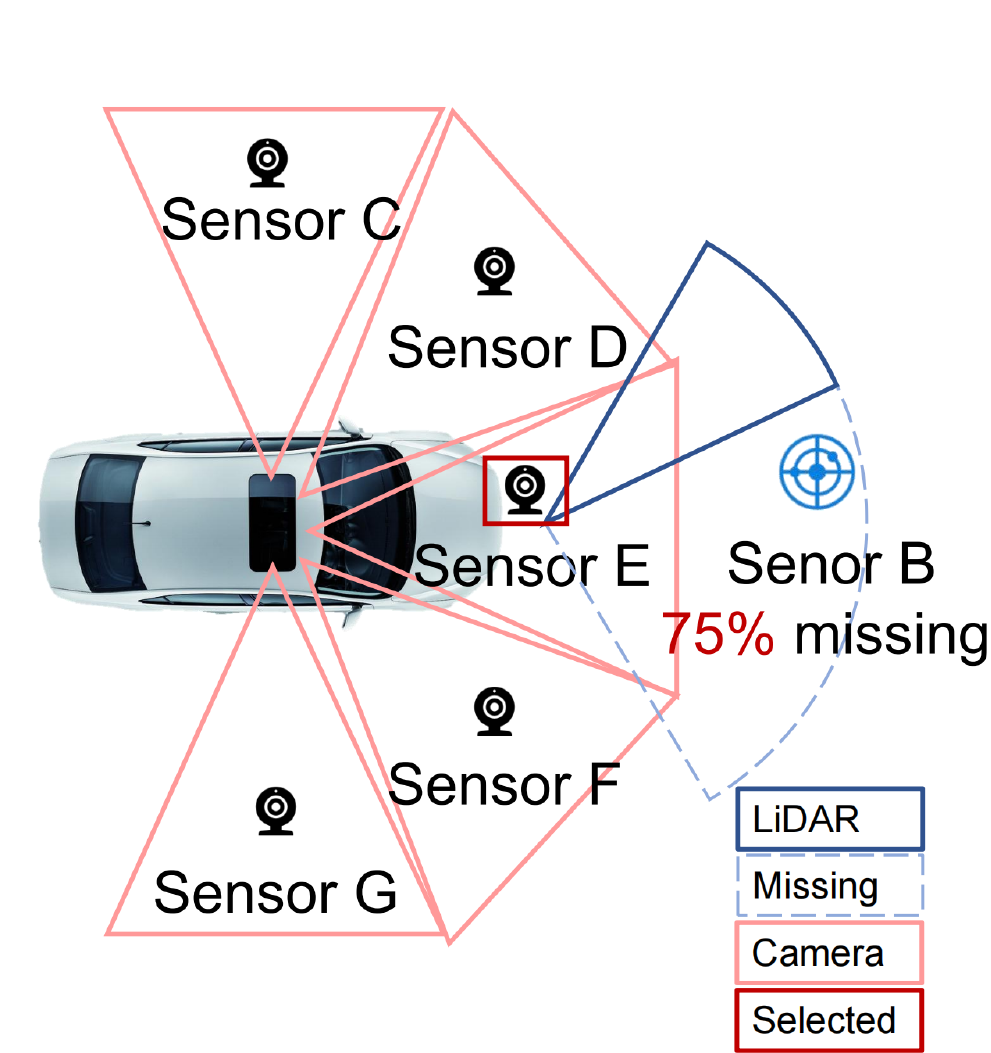}}
    \caption{(a) Top LiDAR emphasizes complementarity and (b) Non-Top LiDAR emphasizes consistency.}
\label{fig:lidar}
\end{figure}

\textit{A. Modality-specific Modality Affinity}.
Different sensors on mobile devices exhibit distinct modality affinities due to their varying view and size, affecting how they interact with data from other modalities, emphasizing consistency and complementarity. 
For instance, consider a vehicle equipped with seven sensors including top and front LiDARs (Sensor $A$ and $B$), and five positioned cameras (sensor C$\sim$G). 
As illustrated in \figref{full}, when the top full-view LiDAR sensor $A$ (slow data with larger size) misses 75\% of its data, selecting data from sensors C$\sim$G enhances \textit{complementarity} and broadens perspectives, thus improving inference accuracy (\ie from 9.6\% to 54.9\%). 
Conversely, \figref{partial} shows that when the front partial-view LiDAR lags 75\% of its data due to network issues, prioritizing data from sensors C$\sim$G for greater \textit{consistency}, rather than \textit{complementarity}, yields better accuracy. 
This is because the partial-view LiDAR provides specific directional data, necessitating supplementary directional data from other cameras to maintain noise-robust results.
This indicates that the types and number of available fast modalities change dynamically in asynchronous systems. Consequently, modality selection for fusion must adapt dynamically to maintain an optimal performance.

\begin{figure*}[t]
	\centering
	\includegraphics[width=0.75\linewidth]{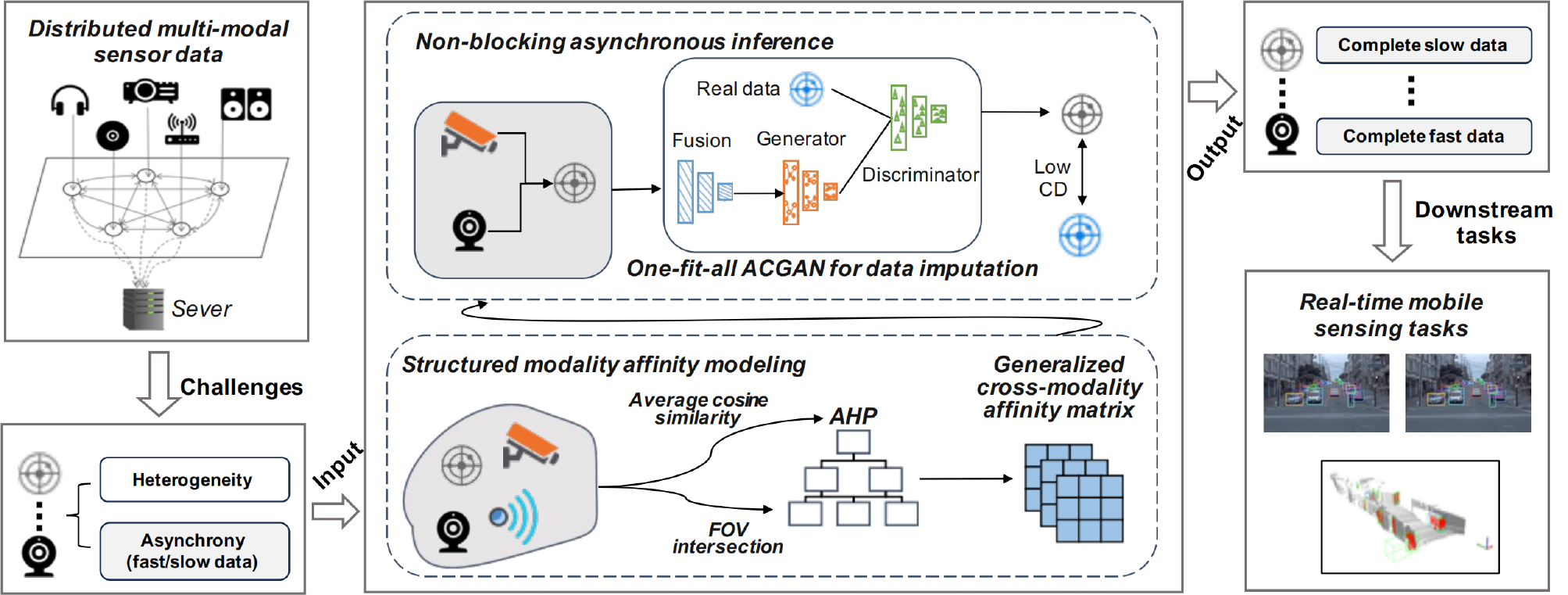}
	\caption{Workflow of \sysname.}
	\label{fig:fw}
\end{figure*}

\begin{figure}[t]
	\centering
	\includegraphics[width=0.7\linewidth]{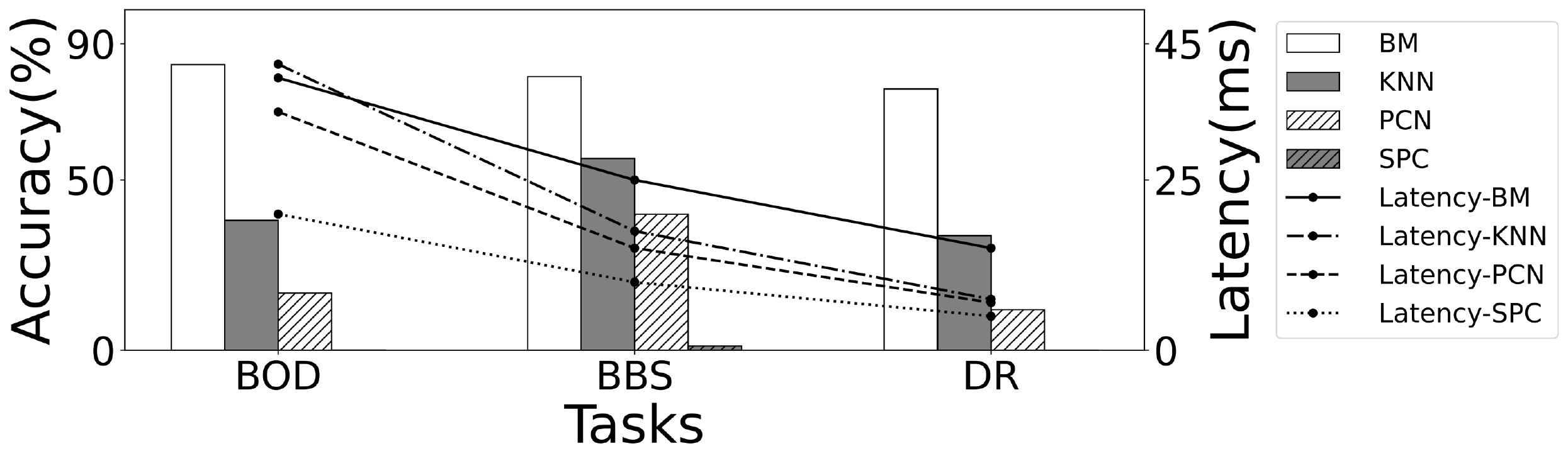}
        \vspace{-2mm}
	\caption{Inference latency and accuracy of BOD, BBS, DR tasks in existing a-/synchronous methods.}
	\label{fig:vt3}
\end{figure}

\textit{B. Subsequent Task-specific Modality Affinity}. 
Even the same multi-modal sensor arrival patterns can exhibit varying modality affinities across different \textit{downstream tasks}. 
As demonstrated in \figref{fig:vt3}, we adopt the same heterogeneity and asynchrony settings, \ie 70ms delays of each frame, corresponding to 50\% missing rate, to fairly test three diverse downstream tasks: \textit{i)  BEV object detection (BOD)}, which involves object recognition from a bird's eye view; ii) \textit{Bounding box segmentation (BBS)}, which uses a rectangular box to represent target objects in images or point clouds for semantic segmentation; \textit{iii) Direction recognition (DR)}, which compares direction similarity between the target and the actual orientation. 
%
The results show that employing a naive synchronous method (\ie BM~\cite{bib:patch:w}) results in significant variations in latency across three tasks, \eg 40ms for BOD, 25ms for BSS, and 15ms for DR task. 
This variance is due to the distinct affinity criteria shaped by downstream tasks, leading to disparities in inference latency especially when data arrives synchronously. 
Such variability is prohibitive in latency-sensitive applications.
%
Moreover, different data imputation techniques for asynchronous fusion (\ie SPC~\cite{bib:mdl:n}, KNN~\cite{bib:knn:c}, PCN~\cite{bib:pcn:Y}) result in even more pronounced variations across three tasks.
For example, when using PCN asychronous method, the inference latency across three tasks are 35ms, 15ms, 7ms, respectively. 
While accuracy are 16.8\%, 39.9\%, and 11.9\%, respectively. 

These observations demonstrate that \textit{dynamics} in mobile sensing patterns, driven by modality asynchrony, heterogeneity, or subsequent tasks, desired \textit{generalized} modality affinity criteria. 
Building this criteria has notable values.

\section{Solution Overview}
\label{sec:solution}
Drawing from the preliminary observations, we introduce \sysname to tackle the challenges of input modality \textit{asynchrony} and \textit{heterogeneity} in mobile distributed multi-modal inference systems (\secref{sec:intro}). 
\figref{fig:fw} illustrates the architecture of \sysname\lsc{(\figref{fig:flow} shows the idea of \sysname)}, comprising two primary modules: Structured Modality Affinity Modeling and Non-blocking \lsc{Asynchronous} Inference.
\textit{First}, the \textit{Structured Modality Affinity Modeling} module, detailed in \secref{sec:matrix}, addresses \textit{Challenge \#1} by constructing a \textit{generalized} cross-modality affinity matrix.
This affinity matrix effectively manages data asynchrony and heterogeneity, enabling dynamic assessments of diverse input modalities.
\textit{Second}, \sysname confronts \textit{Challenge \#2} within the \textit{Non-blocking \lsc{Asynchronous} Inference} module (\secref{sec:inference}). 
\sysname introduces a one-fit-all affinity attention-based conditional GAN (ACGAN), managing various input/output modalities without retraining. 
It ensures real-time, accurate inference via dynamic modality fusion, using a lightweight dynamic programming approach to select optimal affinity sub-graphs based on the affinity matrix.
\lsc{
In real-world deployment, short delays from slow data may make a simple wait-and-process approach sufficient, without requiring modality affinity optimization.
}

\section{Structured Modality Affinity Modeling}
\label{sec:matrix}

\subsection{Generalized Modality Affinity Matrix}
\label{sec:cross}
As valided in \secref{subsec:observation}, distributed multi-modal data in mobile context often vary in views and arrival time, necessitating a \textit{generalized} modality affinity matrix that accommodates diverse \textit{tasks} alongside dynamic input data.

\lsc{
\textbf{Preliminary of the data \textit{consistency} and \textit{complementarity}}.
\textit{Consistency} refers to shared information across modalities, reinforcing fusion accuracy and reliability.
\textit{Complementarity} highlights unique, modality-specific insights.
Both are crucial for robust multimodal fusion, balancing shared and unique data to optimize accuracy and latency.
}

\lsc{\textbf{Limitations of Prior Arts}.
Existing methods have explored data/task affinity matrices~\citep{bib:reid:lu,bib:vip:shi,bib:taskonomy:AZ}; but they mainly focus on \textit{single-view} scenarios~\citep{bib:reid:lu} or \textit{unimodality}~\citep{bib:vip:shi, bib:taskonomy:AZ}, failing to address the \textit{dynamic asynchrony} and \textit{heterogeneity} common in mobile contexts (see \secref{sec:intro}).
For instance, ReID~\citep{bib:reid:lu} uses triplet loss for data affinity in infrared cameras, while a vehicle-road system~\citep{bib:vip:shi} examines bidimensional point clouds, and \citep{bib:taskonomy:AZ} studies task affinity in unimodal visual contexts. 
These methods, tailored to \textit{specific} tasks, are inefficient for real-time mobile context due to the need for extensive customization. 
The key challenge is creating a \textit{generalized} cross-modality affinity matrix to manage heterogeneity, asynchrony, and varied views in the mobile context.
}


In response, we employ t-SNE, a method known for capturing local structures and nonlinear relationships~\citep{bib:t-sne:vd}, to project heterogeneous modalities (of unlimited types) into a high-dimensional feature space.
We then reduce these dimensions while preserving the spatial information by calculating the average cosine similarity at the feature level. 
We utilize cosine similarity at the \textit{feature} level to measure modality affinity because it demonstrates stability despite data heterogeneity (\ie varied modalities) and asynchrony (\ie different rates of missing data).
For instance, the average cosine distance between 3D point cloud and RGB image stands at 0.89192. 
Using features instead of raw data also reduces the size of affinity parameters, facilitating faster processing.
This effectively reflects the consistency and complementarity between diverse input modalities.


Specifically, we utilize t-SNE~\citep{bib:t-sne:vd} to map the Euclidean distances between high-dimensional data points into \textit{joint probabilities} that represent their similarities. 
This method excels at capturing the local structure and nonlinear relationships within the data, which are essential for information-lossless dimensionality reduction.
%
The degree of \textit{similarity} between data points indicates their distance in joint probabilities, where a lower similarity implies a greater distance and higher \textit{complementarity}. 
Thus, we can leverage such cosine similarity as a variable to represent \textit{affinity}. 
In detail, the modality affinity relationship quantified by the joint probability \(p_{ij}\), is calculated as below (line 2 in Algorithm \ref{algorith:affinity}):
\begin{equation}
p_{ij} = \frac{\exp(-\|x_i - x_j\|^2 / 2\sigma_i^2)}{\sum_{k \neq i} \exp(-\|x_k - x_i\|^2 / 2\sigma_i^2)}
\end{equation}
where $\sigma_i$ is the Gaussian variance centered on data point \(x_i\). 
Given the redundancy of sensor data in high-dimensional space, we compress it to a lower dimension to enhance resource efficiency. 
The pair-wise affinity in this reduced space, \(q_{ij}\) (line 5), is computed as follow:
\begin{equation}
q_{ij} = \frac{\exp(-\|y_i - y_j\|^2)}{\sum_{k \neq l} \exp(-\|y_k - y_l\|^2)}
\end{equation}

To minimize information loss during the above dimensionality reduction process, we minimize the Kullback-Leibler (KL)~\citep{bib:kl:k} divergence,
a measure commonly used to assess the information loss between two probability distributions.
Specifically, we minimize the KL divergence between
$p_{ij}$ and $q_{ij}$, which represent the high- and low-dimensional distributions of data point pairs, respectively.
The optimization technique used in t-SNE is gradient descent, effectively minimizing the total KL divergence across data point pairs. 
The cost function $C$ for this optimization is as follows:
\begin{equation}
C = KL(P||Q) = \sum_i \sum_j p_{ij} \log \frac{p_{ij}}{q_{ij}}.
\end{equation}
where $P$ is the affinity value in the high-dimensional space, $Q$ is the corresponding affinity value in the low-dimensional space. 
The gradient formula is presented below:
\begin{equation}
\frac{\partial C}{\partial y_i} = 4 \sum_j (p_{ij} - q_{ij})(y_i - y_j).
\end{equation}

Thus we have iteration as follows:
\begin{equation}
\gamma^{(t)} = \gamma^{(t-1)} + Adam( \frac{\partial C}{\partial \gamma} )
\end{equation}
where \(\gamma^{(t)}\) is the solution at iteration $t$, $Adam$ represents Adam optimizer~\citep{bib:adam:k}.
Upon obtaining the low-dimensional mapping of each modality at time \(t\), we can readily calculate the average cosine similarity for each pair-wise modalities using t-SNE, as outlined in Algorithm \ref{algorith:affinity}.


We demonstrate the above process using Waymo~\citep{bib:waymo:s} dataset, real-world autonomous driving data that includes three sensors: the front LiDAR, front camera, and left-side camera. 
They experience asynchrony due to varying data volumes. 
We perform t-SNE dimensionality reduction and visualization for the front LiDAR paired with the front camera, and separately for the front LiDAR with the left-side camera, with color-coded mappings to distinguish sensor data. 
The mapping of front LiDAR and front camera data shows a predominance of blue points—indicative of the LiDAR's superior spatial data capture—alongside a common trend with green points. 
\lsc{
Typically, the threshold is set to the median in samples, which for the Waymo dataset is experimentally set around 0.06. 
Values above this indicate high consistency, while those below means high complementarity.} 
The cosine similarity of 0.10808 between these sensors, as shown in Fig 5a, highlights their significant consistency and overlapping fields of view.
Conversely, Fig 5b illustrates the mapping between the front LiDAR and the left-side camera, revealing a lower correlation with a cosine similarity of 0.03346. 
This indicates that these modalities provide complementary information due to differing views.

\begin{figure}[t]
    \centering 
    \subfloat[]{\label{fig:ff}
    \includegraphics[width=0.32\linewidth]{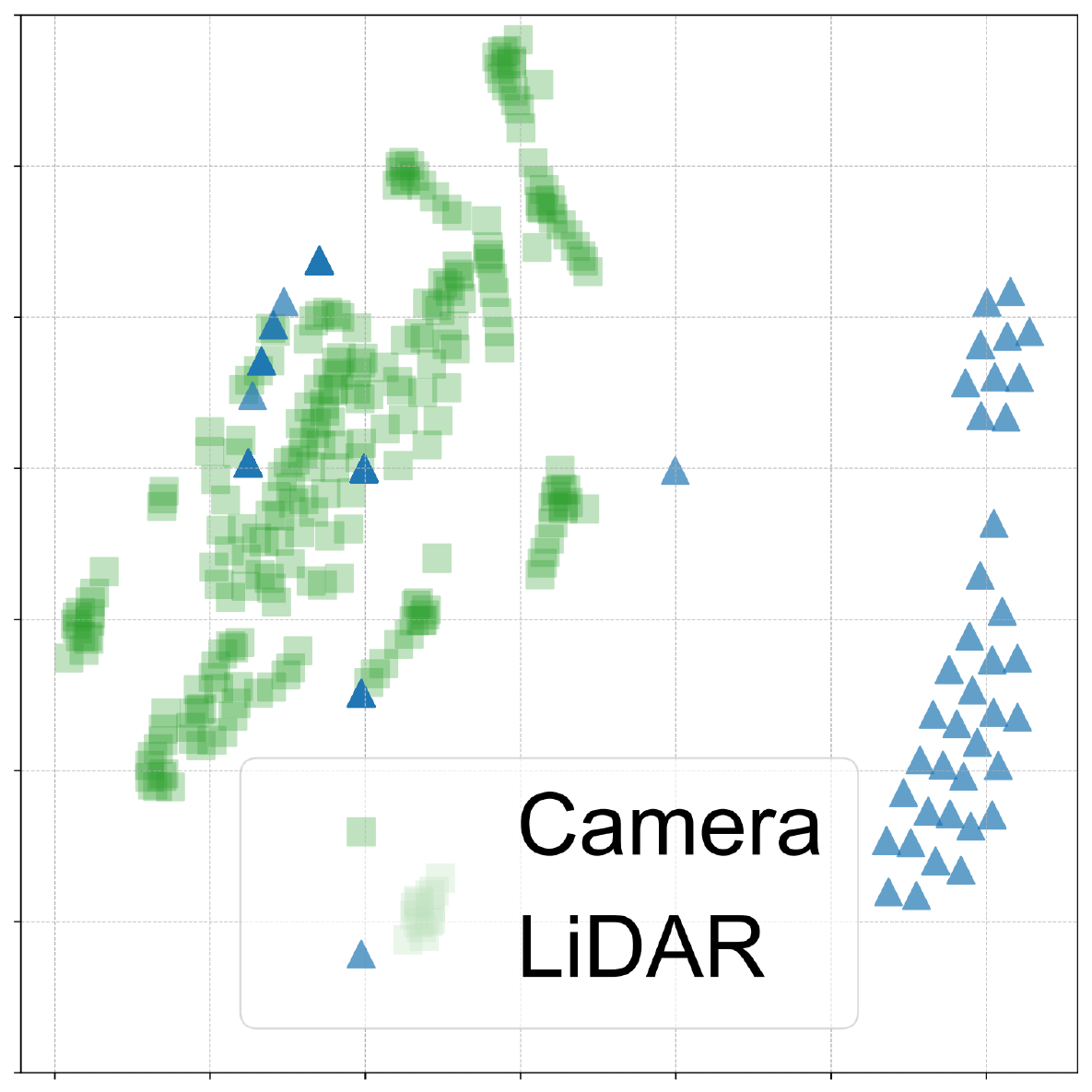}}
    \hspace{18mm}
     \subfloat[]{\label{fig:fl}
    \includegraphics[width=0.32\linewidth]{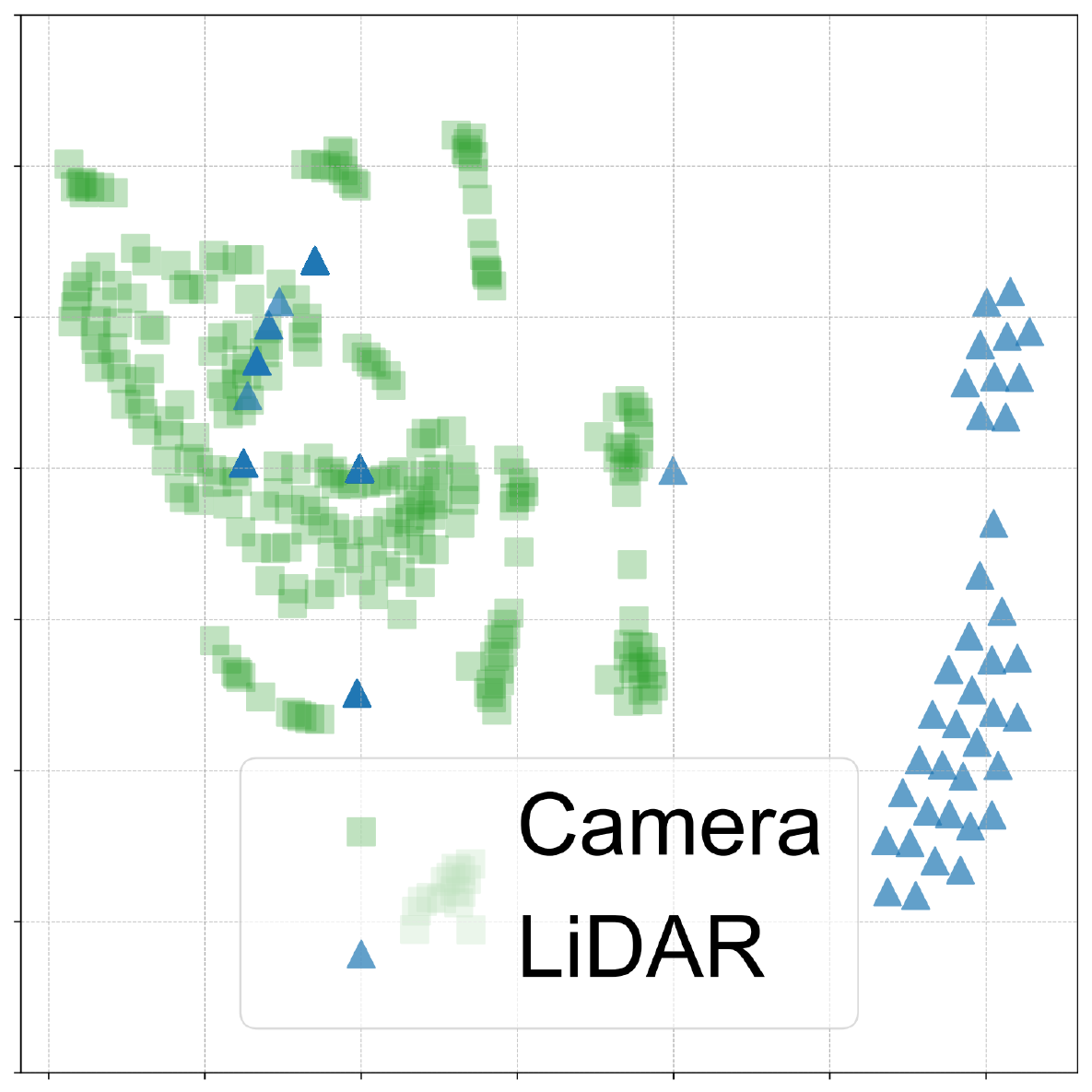}}
    \vspace{-2mm}
    \caption{Visualization of (a) front LiDAR $\&$ front camera features and (b) front LiDAR $\&$ left camera features.}
\label{fig:ffl}
\end{figure}

\begin{algorithm}[t]
\caption{Generalized Cross-modal Affinity Matrix}
\label{algorith:affinity}
\begin{algorithmic}[1]
\Require data set $X_k = \{x_{k1}, x_{k2}, \dots, x_{kn}\}$, number of iterations $T$, average cosine similarity $S$.
\Ensure low-dimensional data representation $\gamma_k = \{y_{k1}, y_{k2}, \dots, y_{kn}\}$.
\For{$k=0$ to $1$}
\State Compute pairwise affinities $p_{ij}$ (using Equation 1)
\State Sample initial solution $\gamma_k^{(0)} = \{y_{k1}^{(0)}, y_{k2}^{(0)}, \dots, y_{kn}^{(0)}\}$ from $\mathcal{N}(0, 10^{-4})$
\For{$t = 1$ to $T$}
    \State Compute low-dimensional affinities $q_{ij}$ (using Equ. 2)
    \State Compute gradient $\frac{\partial C}{\partial \mathcal{Y}}$ (using Equ. 4)
    \State Set $\gamma_k^{(t)} = \gamma_k^{(t-1)} + Adam( \frac{\partial C}{\partial \gamma_k} )$
\EndFor
\EndFor
\State $S$ = Cos( $\gamma_0^{(t)},\gamma_1^{(t)}$) 
\end{algorithmic}
\end{algorithm}

\begin{figure}[t]
	\centering
	\includegraphics[width=0.7\linewidth]{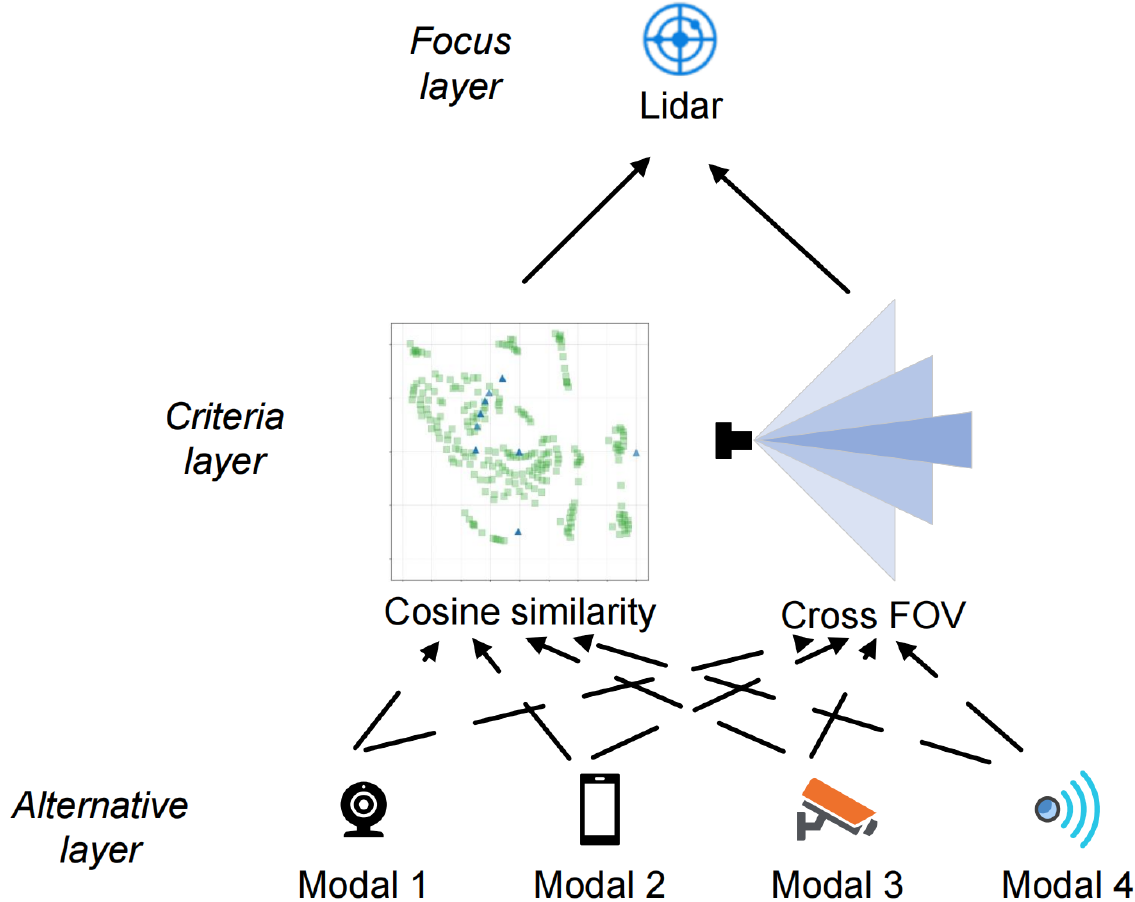}
 \vspace{-2mm}
	\caption{Affinity matrix normalization using AHP.}
	\label{AHP}
\end{figure}

\subsection{Normalization using Analytic Hierarchy Process (AHP)}
\label{matrix}
In mobile systems equipped with multiple modality types and volumes, integrating the affinity measure between each pair of modalities into a unified framework is intractable.

Existing methods primarily model features within the same dimension~\citep{bib:taskonomy:AZ,bib:reid:lu} or targeting specific characteristics like locations, semantics, sizes, and spatial distribution~\citep{bib:dnn:h,bib:ml:s,bib:vip:shi}.
For instance, ReID~\citep{bib:reid:lu} employs normalized Euclidean distance for affinity matrix elements within the same dimension. 
VIPS~\citep{bib:vip:shi} examines node-to-node and edge-to-edge relationships in point cloud data. 
However, they do not effectively extend to multi-modal problems. 
Normalization techniques including Quality-aware Multi-modal Fusion~\citep{bib:prov:z} and taskonomy~\citep{bib:taskonomy:AZ} also exhibit limitations in addressing the system asynchrony (\ie varying missing rates of partial data) and heterogeneity (\ie modality type) challenges of distributed modality.

To effectively manage the diversity and dynamics of multi-modal fusion, we utilize the Analytic Hierarchy Process (AHP) to \textit{normalize} affinity values.
Normalization in this context means adjusting values measured on different scales to a notionally common scale.
Although obtaining pair-wise modalities' affinity using Algorithm \ref{algorith:affinity} is a straightforward procedure, applying this algorithm globally (across all data and modalities) and incorporating every single pair-wise modalities' affinity into a unified framework transforms the problem into a complex \textit{graph programming} issue.
\lsc{
Since the problem reduces to the vertex cover problem~\citep{bib:np:D}, it is NP-hard.
In particular, we define the problem as $Q$, with a reduction algorithm $f$ such that $Q = f(G(v_i, k))$, where $G(v_i, k)$ is a solution to the vertex cover problem. 
Algorithm $f$ treats modalities as vertices $v$ and their affinities as weighted edges in a graph. 
Setting $k=1$ identifies the global optimal solution. 
Finding a subset of vertices based on weights has a time complexity of $O(n)$, meaning $f$ runs in $O(n)$ time. 
}

We observe that the complexity of the problem and the fact that sub-optimization is more resource-efficient, we are not aiming for a globally optimal solution. 
Instead, we can start with slow data for planning, which reduces the complex problem to a multi-objective programming problem~\citep{bib:multiob:e} that is more suitable for AHP.
In particular, our designed AHP quantitatively standardizes modality affinity and replaces subjective expert judgments with \textit{empirical data}, thus reducing the controversies often associated with traditional AHP.
The affinity matrix in our AHP has three layers: the \textit{Focus}, \textit{Criteria}, and \textit{Alternative} layers. 
The \textit{Focus} layer prioritizes sensors with lower data generation rates or higher vulnerability to lagging, while the \textit{Alternatives} layer includes sensors that transmit data to the terminal first. 
The \textit{Criteria} layer considers two main factors: the average cosine similarity from the t-SNE visualization and the Field Of View (FOV) intersection between different sensors. 
Initially, we intend to use mean cosine similarity as the sole affinity indicator; however, to mitigate potential biases in practice, we incorporate FOV intersection as an auxiliary measure.

\begin{table}[t]
\centering
\tiny
\caption{The scale of importance between sensors.}
\begin{tabular}{cllX}
\toprule
Intensity of importance $X_i$($X=A,B$) & \multicolumn{2}{c}{Definition} \\
on an absolute scale & & \\
\midrule
$x_{ij}$ = 1 & \multicolumn{2}{l}{Equal importance between sensor i and j} \\
$x_{ij}$ = 3 & \multicolumn{2}{l}{Moderate importance of sensor i over sensor j} \\
$x_{ij}$ = 5 & \multicolumn{2}{l}{Essential or strong importance of sensor i over sensor j} \\
$x_{ij}$ = 7 & \multicolumn{2}{l}{Very strong importance of sensor i over sensor j} \\
$x_{ij}$ = 9 & \multicolumn{2}{l}{Extreme importance of sensor i over sensor j} \\
$x_{ij}$ = 2, 4, 6, 8 & \multicolumn{2}{l}{Intermediate values between the two adjacent judgments} \\
Reciprocals & \multicolumn{2}{l}{If there is an established importance between \(i\) and \(j\),}\\
& \multicolumn{2}{l}{then the importance of \(j\) to \(i\) is the reciprocal of the $x_{ij}$.} \\
\bottomrule
\label{table:fundamental}
\end{tabular}
\end{table}

To illustrate how the Analytic Hierarchy Process (AHP) is used to establish an affinity matrix, let's consider a simple example. 
Our AHP process consists of three layers, resulting in two layers of consistency matrices (conformity test matrices) named A and $\{B_1, B_2\}$. 
Within matrix A, the element \(a_{ij}\) represents the relative importance of each pairwise criterion contrast in the criterion layer. 
The scaling method for \(a_{ij}\) is detailed in \tabref{table:fundamental}.
In this example, matrix A is straightforward, consisting of two rows and two columns. 
\(C_1\) represents average cosine similarity and \(C_2\) represents the intersection on the Field Of View (FOV). 
We assign \(a_{12} = 7\) and \(a_{21} = \frac{1}{7}\), emphasizing that FOV serves as an auxiliary adjustment rather than a primary measure, reflecting its lesser importance in this context. 
\lsc{\(a_{ij}\) varies with the data signal-to-noise ratio. 
Increasing \(a_{21}\) and decreasing \(a_{12}\) can help correct bias.}

\[
A=\begin{array}{c}
\quad\quad{C_1} \quad {C_2} \\
\begin{array}{c}
{C_1} \\
{C_2} \\
\end{array}
\left[\begin{array}{cccc}
1 & 7 \\
\frac{1}{7} & 1 \\
\end{array}\right]
\end{array}
\]
In the matrix \(B_i\), the element \(b_{ij}\) represents the importance of sensor \(i\) relative to sensor \(j\) under criterion \(C_i\). 
This importance is derived by standardizing the average cosine or FOV values of each sensor in t-SNE, thus excluding subjective expert judgment. 
For example, with one Focus sensor and three Alternative sensors \(\{sensor_1, sensor_2, sensor_3\}\), the matrix \(B_i\) is a $3\times 3$ matrix.

\[
B_1=\begin{array}{c}
\begin{array}{c}
\end{array}
\left[\begin{array}{cccc}
1 & 2 & 5 \\
\frac{1}{2} & 1 & 2\\
\frac{1}{5} & \frac{1}{2} & 1\\
\end{array}\right]
\end{array}
B_2=\begin{array}{c}
\begin{array}{c}
\end{array}
\left[\begin{array}{cccc}
1 & \frac{1}{3} & \frac{1}{8} \\
3 & 1 & \frac{1}{3}\\
8 & 3 & 1\\
\end{array}\right]
\end{array}
\]
where \(A\) and \(\{B_1, B_2\}\) all pass the consistency test, their eigenvalues can be approximated by the arithmetic mean of their column vectors, thus obtaining the eigenvectors.

\[
W_1=\begin{array}{c}
\begin{array}{c}
\end{array}
\left[\begin{array}{cccc}
0.875  \\
0.125 \\
\end{array}\right]
\end{array}
W_2=\begin{array}{c}
{B_1} \quad\quad {B_2} \\
\begin{array}{c}
\end{array}
\left[\begin{array}{cccc}
0.594 & 0.082 \\
0.277 & 0.236 \\
0.129 & 0.682 \\
\end{array}\right]
\end{array}
\]
Then,
\[
W=W_1\cdot W_2=\begin{array}{c}
\begin{array}{c}
\end{array}
\left[\begin{array}{cccc}
0.53  \\
0.272 \\
0.198 \\
\end{array}\right]
\end{array}
\]
According to values in \(W\), for the Focus sensor, \(B_1\) has the greatest affinity, followed by \(B_2\), with \(B_3\) being the least.
This process yields a vector \(W\) for each sensor.
By combining these vectors, we form the normalized affinity matrix.

\section{Non-blocking \lsc{Asynchronous} Inference}
\label{sec:inference}


This subsection then explores the adaptive fusion of asynchronous and heterogeneous input via imputation, ensuring non-blocking inference without compromising accuracy.



\subsection{One-fit-all Data Imputation}
\label{gan}

\textbf{Why conditional GAN (CGAN)}.
A conditional generative adversarial network (CGAN) becomes viable for \sysname to predict missing or delayed input data across different modalities. 
CGANs excel in data imputation due to three key advantages~:
\textit{i)} CGANs incorporate conditional inputs that govern the attributes of the generated data, aligning it more closely with the specific target distribution~\cite{bib:cgan1:m}. 
\textit{ii)} CGANs excel in managing high-dimensional data and complex modalities. 
This is particularly beneficial for serial data streams~\cite{bib:cgan2:l}.
\textit{iii)} Shallow imputation methods treat each missing data independently, often leading to inconsistencies~\cite{bib:cgan3:l}. 
While CGANs learn the intrinsic structure and correlations within data, ensuring imputations are highly consistent with the actual distributed multi-modal distribution.

\textbf{Preliminary of CGAN}. 
A CGAN consists of two interlinked blocks: the generator \(G\) and the discriminator \(D\). 
The generator learns to create data that mimics the target distribution from a latent space, while the discriminator evaluates the authenticity of samples generated by \(G\), providing feedback that improves \(G\)'s outputs. 
These models engage in dynamic competition, continually adjusting to better simulate or identify real data. 
When conditional generation is required, \(G\) can incorporate specific data conditions to tailor the output, utilizing inputs from various data streams to enhance the completeness and relevance of its predictions.

However, due to \textit{dynamic asynchrony} and \textit{heterogeneity} in mobile sensing patterns, directly applying conditional GAN to a distributed multi-modal system is often ineffective. 
Traditional CGANs typically assume \textit{synchronous} inputs with fixed input volume and type, which is not the case in mobile distributed systems where, for example, two types of LiDAR require different camera characteristics for $G$. 

\begin{figure}[t]
	\centering
	\includegraphics[width=0.9\linewidth]{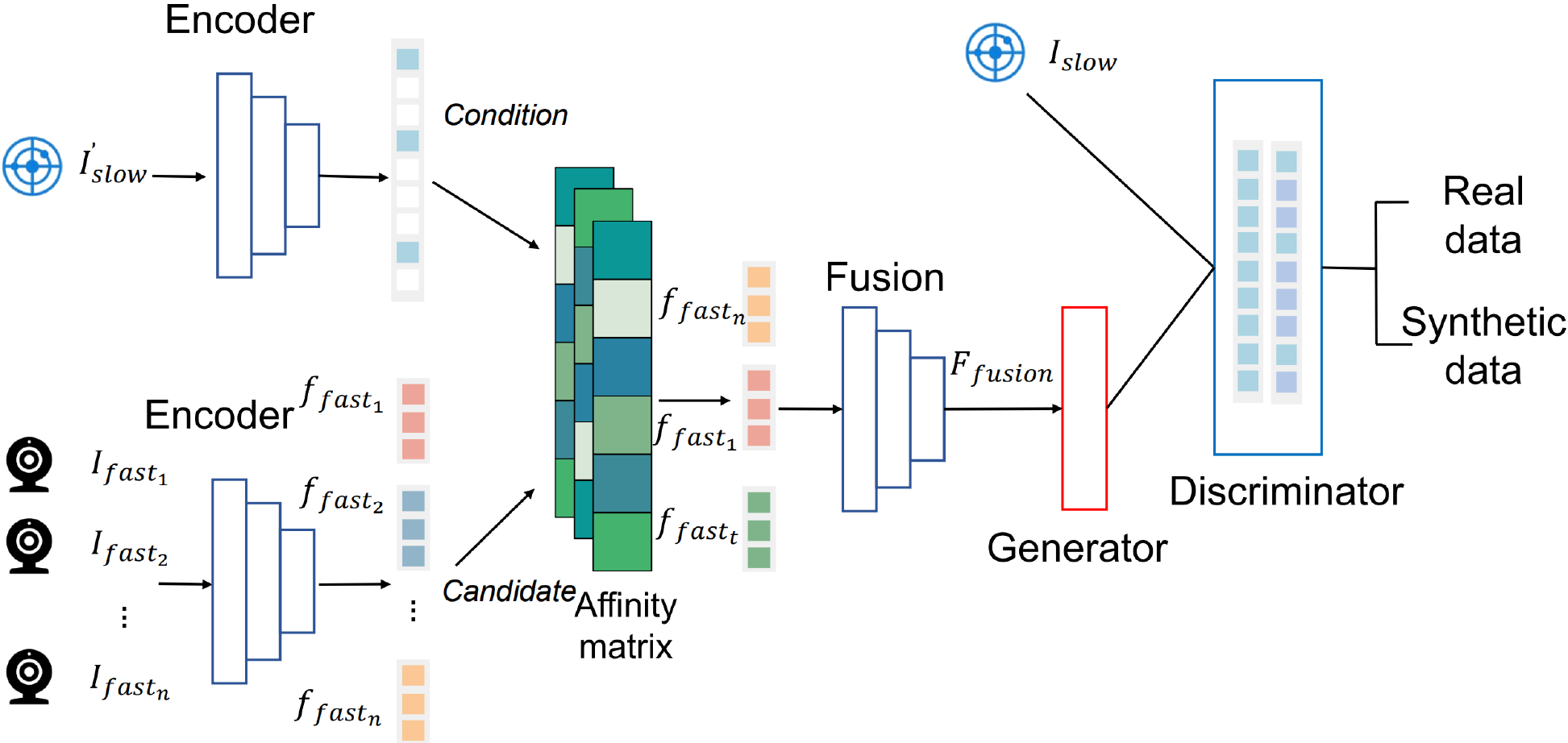}
	\caption{ACGAN-based non-blocking inference.}
	\label{fig:gan}
\end{figure}

To tackle these challenges, we present an affinity attention-based conditional GAN (ACGAN), which selectively incorporates diverse amounts and types of modality as inputs.
\lsc{
Unlike~\cite{bib:lls:L}, which only accepts input from modality $I_{fast_1}$ to generate slow data $I_{LiDAR}'$, our approach is one-fit-all, capable of accepting various inputs (\eg any combination of inputs $I_{fast_1}$, $I_{fast_2}$, or $I_{fast_3}$) to generate data for slow data $I_{LiDAR}'$.
}
As depicted in \figref{fig:gan}, the ACGAN integrates an attention-based fusion mechanism. 
It computes an attention matrix through \textit{affinity calculation} and utilizes this matrix to perform a weighted fusion of features. 
This process selectively fuses appropriate fast and slow data, optimizing data fusion.

Specifically, as shown in \figref{fig:gan}, the ACGAN has an attention-based fusion, which generates an attention matrix through \textit{affinity calculation} and uses the matrix to perform the weighted fusion of features to select appropriate fast data flows and slow data flows for fusion.
Using LiDAR and cameras as examples, we extract camera features $F_{camera_1}$, $F_{camera_2}$, ..., $F_{camera_i}$ from the affinity matrix \(h\) and LiDAR features $F_{LiDAR}$ from incomplete data $I_{LiDAR}'$ through a fusion layer \(g\).
This results in combined features $F_{fusion}$, which serve as inputs for the generator \(G\) to generate synthetic, complete LiDAR data $I_{LiDAR}^*$, replicating the characteristics of the full LiDAR dataset $I_{LiDAR}$.
 
\textbf{Training of ACGAN}. 
The training objective for ACGAN consists of two components: the conditional ACGAN loss \(\mathcal{L}_{CGAN}\), and a loss function for slow data stream characteristics \(\mathcal{L}_{*}\), which measures the discrepancy between modalities.
Inputs to the generator \(G\) include complete fast data stream \{\textbf{\(I_{fast_1}\)}, \textbf{\(I_{fast_2}\)}, ..., \textbf{\(I_{fast_n}\)}\} and the incomplete slow data stream \textbf{\(I_{slow}^{'}\)}.
This setup enables \(G\) to generate the synthetic slow data \textbf{\(I_{slow}^{*}\)}. 
The ACGAN loss function is formulated as follows:
\begin{equation}
\begin{split}
&\{\textbf{$F_{fast_1}$},\textbf{$F_{fast_2}$},...,\textbf{$F_{fast_n}$}\}=\\&\textbf{$f_{fast}$}(\{\textbf{$I_{fast_1}$},\textbf{$I_{fast_2}$},...,\textbf{$I_{fastn}$}\})
\end{split}
\end{equation}
\begin{equation}
\textbf{$F_{slow}$}=\textbf{$f_{slowR}$}(\textbf{$I_{slow}^{'}$})
\end{equation}
\begin{equation}
\textbf{$F_{fusion}$}=
g(h(\{\textbf{$F_{fast_1}$},...,\textbf{$F_{fast_n}$}\}),\textbf{$F_{slow}$})
\end{equation}
\begin{equation}
\textbf{$I_{slow}^{*}$}=G(\textbf{$F_{fusion}$})
\end{equation}
\begin{equation}
\begin{split}
&   
\mathcal{L}_{CGAN}=\mathbb{E}_{\textbf{$I_{slow}$}}[logD(\textbf{$I_{slow}$})]\\
&+\mathbb{E}_{\textbf{$I_{slow}^{*}$}}[log(1-D(\textbf{$I_{slow}^{*}$}))]
\end{split}
\end{equation}
where the generator \(G\) seeks to minimize Equ. 10, while the adversarial discriminator \(D\) strives to maximize it. 


For example, in the task of 3D object detection for autonomous driving assistance, the camera data is always the fast stream, whereas the LiDAR data is slow. 
Due to the characteristic nature of point cloud produced by LiDAR, we opt for the Chamfer distance loss \(\mathcal{L}_{CD}\)(Chamfer distance loss) between real point cloud \(X\) and synthetic point cloud \(Y\) as \(\mathcal{L}_{*}\) which is expressed as follows:
\begin{equation}
X=\{x_i\}_{i=1}^N Y=\{y_i\}_{i=1}^M,X=\textbf{$I_{LiDAR}^{*}$} 
 Y=\textbf{$I_{LiDAR}$}
\end{equation}
\begin{equation}
\mathcal{L}_{CD}=\sum_{x \in X} \min_{y \in Y} \|x - y\|_2^2 + \sum_{y \in Y} \min_{x \in X} \|y - x\|_2^2
\end{equation}
For each point \(x\) in point cloud \(X\), we identify the nearest point \(y\) in point cloud \(Y\) and compute the squared Euclidean distance between them, summing these values for all points in $X$. 
Similarly, for each point \(y\) in \(Y\), we find the nearest point \(x\) in  \(X\), calculate the squared Euclidean distance between them, and sum these values for all points in $Y$.
Thus, our final objective function becomes:
\begin{equation}
\mathcal{L}=arg\,\underset{G}{\min} \, \underset{D}{\max}\,\mathcal{L}_{CGAN}+\lambda_{CD}\mathcal{L}_{CD}
\end{equation}
where $\lambda_{CD}$ is a hyper-parameter to tune the data fitting and model complexity.
\lsc{
As a separate note, in autonomous driving, LiDAR data is generally slower than camera data. For instance, any combination of cameras $I_{fast_1}$, $I_{fast_2}$, and $I_{fast_3}$ (with asynchrony) can generate LiDAR data $I_{LiDAR}'$, with the loss function incorporating point cloud. 
If the downstream task is fixed, asynchronous data can fill in for imputing the slower LiDAR without retraining. 
The loss function can also be adjusted for other modalities in different tasks.
}

\subsection{Affinity-aware Sub-graph Selection}
\label{graph}
Existing data imputation methods~\cite{bib:gain:y,bib:miwae:m} predefine the type and number of modalities for data fusion and imputation, assuming fixed consistency and complementarity based on learned sensing patterns. 
However, these methods are unsuitable for mobile systems characterized by dynamics~\cite{bib:lls:L}. 
\lsc{
Studies~\cite{bib:incomplete:w,bib:less:d} reveal potential drawbacks. Specifically, \cite{bib:incomplete:w} shows that adding more modalities without \textit{key} ones can introduce noise and redundancy, degrading performance. 
Similarly, \cite{bib:less:d} reports that not all modalities contribute positively, with some introducing unnecessary noise. 
To address this, \sysname selectively chooses modalities that enhance fusion for better imputation (as we will show in \secref{val}.)
}

For example, \sysname uses real-world collected data in autonomous driving assistance, utilizing two LiDAR sensors and five cameras. 
The LiDARs are strategically positioned at the top and front, while the cameras capture images from various perspectives, including front, front-left, left, front-right and right. 
These sensors correspond to sensors A$\sim$G in \figref{full} and \figref{partial}.
Specifically, for the top LiDAR, which produces 360° point clouds in the slower data stream, we set a missing rate of 75\% to mimic the delay typically observed in slow data streams, as shown in \figref{full}. 
Given the broad sensing view of sensor A, its criteria for selecting fusion modalities should prioritize \textit{complementarity} over \textit{consistency}. 
Thus, fast data in sensor A with smaller cosine similarity are preferred for fusion to match the slow data stream, as detailed in \secref{full}. 
Here, smaller cosine similarity, indicating larger average cosine distances, denotes a high level of \textit{complementarity} between heterogeneous modalities. 
With a 75\% missing rate of front LiDAR sensor B, selecting RGB images from other sensors with higher cosine similarity provides limited benefits in fusion accuracy.
Similarly, for point clouds generated by other LiDAR sensors in the slow stream, we also maintain a 75\% missing rate.

We note that the imputation may \textit{vary} according to different data arrival patterns. 
The problem can be described as a \textit{decision graph}: given a set of heterogeneous modalities \( M = \{m_1, m_2, \ldots, m_N\} \) as input, 
output the map-wise complementarity or consistency decision $m_{fi}\rightarrow m_{sj}$, where \lsc{$m_{sj}$ }is the slow modality and $m_{fi}$ is the fast modality. It can be formulated as $m_{fi}\rightarrow m_{sj}$ = $\text{F}(M)$.
For $\text{F}(M)$, we assess complementarity and consistency using the intersection over union between modalities. 
The intersection over union between modalities is defined as:
\begin{equation}
    v_{ij} =  \frac{\text{FOV}_{m_{sj}} \cap \text{FOV}_{m_{fi}}}{\text{FOV}_{m_{sj}} \cup \text{FOV}_{m_{fi}} }
\end{equation}

And if \(v_{ij}\) $\not=$ 0, \(v_{ij} \in V_j\). For each \(m_{sj} \in S\), we set:
\begin{equation}
 m_{fi}\rightarrow m_{sj} = 
 \begin{cases}
Consistency\ ,  &\exists v_{ij}=0\\
Complementarity, &\exists\mkern-10.5mu/ \ v_{ij}=0
\end{cases}
\end{equation}

For \(m_{sj}\), its' criteria of affinity has settle down, \( a_{ij} \) is the affinity between \( m_{fi} \) and \( m_{sj} \),
Then we have:
\begin{equation}
    A=\sum_{}^{j} \sum_{i=1}^{k}a_{ij}
\end{equation}
Where the value of \(k\) as shown below:
\begin{equation}
k=\lfloor |V_j| \times r \rfloor
\end{equation}

Where \(r\) is the data missing rate of partial (slow) data and $|V_j|$ is the number of intersecting FOV sensors. We need to discuss for value \(k\):
\begin{equation}
k=
\begin{cases}
k, k \not= 0 \\
1, k = 0 \ |V_j| \not= 0\\
0, k=0  \ |V_j| =0
\end{cases}
\end{equation}
By maximizing \(A\) (Equ. 16), we obtain the 
$m_{fi}\rightarrow m_{sj}$ map. 

Based on the arrival speed and availability of input data streams, \sysname adaptively determines the selection of \textit{affinity sub-graph} for data fusion by maximizing \(A\), \ie which modality and modality \( S \) to fuse, and which are the target tasks \( F\) using the selected affinity sub-graph, to maximize the imputation effect.
In particular, based on the affinity matrix as discussed in \secref{matrix}, the sub-graph selection criteria for runtime fusion are dynamically determined by the specific requirements of the downstream task, influenced by the asynchrony and heterogeneity of input data. 
\section{Experiments}
\label{exp}
\subsection{Setups}

\textbf{Implementation.} 
AdaFlow is implemented using Python 3.9 and PyTorch 1.11 on a server equipped with an RTX3090 GPU, Intel(R) Xeon(R) Gold 6133 CPU @ 2.50GHz, and 256GB RAM. 
For distributed multi-modal data on mobile devices, we realize both simulation with real-world datasets and actual collections. 
We also simulated the bandwidth of the real car Internet, which is 100Mbps.
In the implementation of the affinity matrix (\secref{matrix} ), we designate LiDAR as the primary modality and cameras as secondary modalities. For the ACGAN architecture (\secref{gan}), we employ two feature-capturing networks to process the raw data. 
These networks, via the affinity matrix, select the appropriate modalities for fusion. 
The resultant fusion features are then inputted into the generator to synthesize the primary modality data.

\textbf{Tasks, Datasets, and Model.} We experiment with 4 real-world distributed multi-modal applications.
They are widely used to evaluate multi-modal fusion performance~\cite{bib:nuscenses:ca,bib:kitti:g,bib:da:m}.
\begin{itemize}
\item \textbf{BEVFusion ($T_{1}$)} is a real-world multi-modal 3D object recognition task using a bird's-eye view.
The \textbf{dataset} (nuScenes ~\cite{bib:nuscenses:ca}) contains 2000 training frames and 81 testing frames, each featuring 6 camera views (fast data) and 1 LiDAR views (slow data).
The \textbf{model} for BEVFusion extracts a bird's-eye view (BEV) from both the camera and LiDAR branches.

\item \textbf{PointPillars ($T_{2}$)} is another real-world 3D object recognition model. 
The \textbf{dataset} from KITTI~\cite{bib:kitti:g} includes 2500 frames for training, and 81 frames for testing, all collected from actual autonomous driving systems, with each frame featuring 4 camera views (fast data) and 1 LiDAR view (slow data)
The \textbf{model} projects both camera and LiDAR data into a unified space and merges multiple views.
\item 
\textbf{LLM-based driver behavior recognition ($T_{3}$)} employs \textbf{Large Language Model (LLM)}~\cite{bib:one:h} to categorize driver behaviors into 12 classes using the Drive\&Act dataset~\cite{bib:da:m}, which comprises 6 infrared camera views (fast data) and 1 standard camera view (slow data).
\item 
\textbf{LLM-based event recognition (\(T_4\))} employes \textbf{LLM} for audio-visual event recognition across 28 categories using the AVE dataset~\cite{bib:ave:t} which contains audio (fast data) and video (slow data).


\end{itemize}


\textbf{Baselines.} We adopt five a-/synchronous fusion methods as comparison baselines to validate the affinity matrix.
\begin{itemize}
\item \textbf{Synchronous fusion} (blocking mechanism, BM)~\cite{bib:patch:w}: 
the inference pipeline waits until all input data arrive. 
It provides an upper-bound accuracy.
\item \textbf{Asynchronous fusion} (non-blocking mechanism):
\begin{itemize}
    \item Sparse Point Cloud (SPC)~\cite{bib:mdl:n}: discards the slow data for inference which is the fastest method but exhibits the lowest accuracy.
    \item PCN~\cite{bib:pcn:Y}: is a DL model to impute 3D point clouds. When the point cloud is missing due to data asynchrony, we use it to complete the point cloud.
    \item KNN~\cite{bib:knn:c}: \lsc{is a traditional imputation method, which classifies an input by majority class among its $K$ nearest neighbors to selectively} discard data.
    \item Traditional CGAN~\cite{bib:cgan:m}: has the same imputation network structure as \sysname, but selects fixed or random fast data to impute the slow data. 
    We use it to prove the effectiveness of the affinity matrix. 
    
\end{itemize}
\end{itemize}

\subsection{Performance Comparison}
We evaluate the inference accuracy and latency of AdaFlow against four baseline methods (BM, SPC, PCN, and KNN) across BEVFusion ($T_1$) and PointPillars ($T_2$) tasks under various missing rates of slow data, which simulate data asynchrony. 
For example, a 25\% missing rate of slow data represents a delay of 26ms in dataset nuScenses with 100Mbps.

\begin{figure}[h]
  \centering
  \subfloat[BEVFusion, missing 50\% LiDAR.]{
    \includegraphics[height=0.1\textwidth]{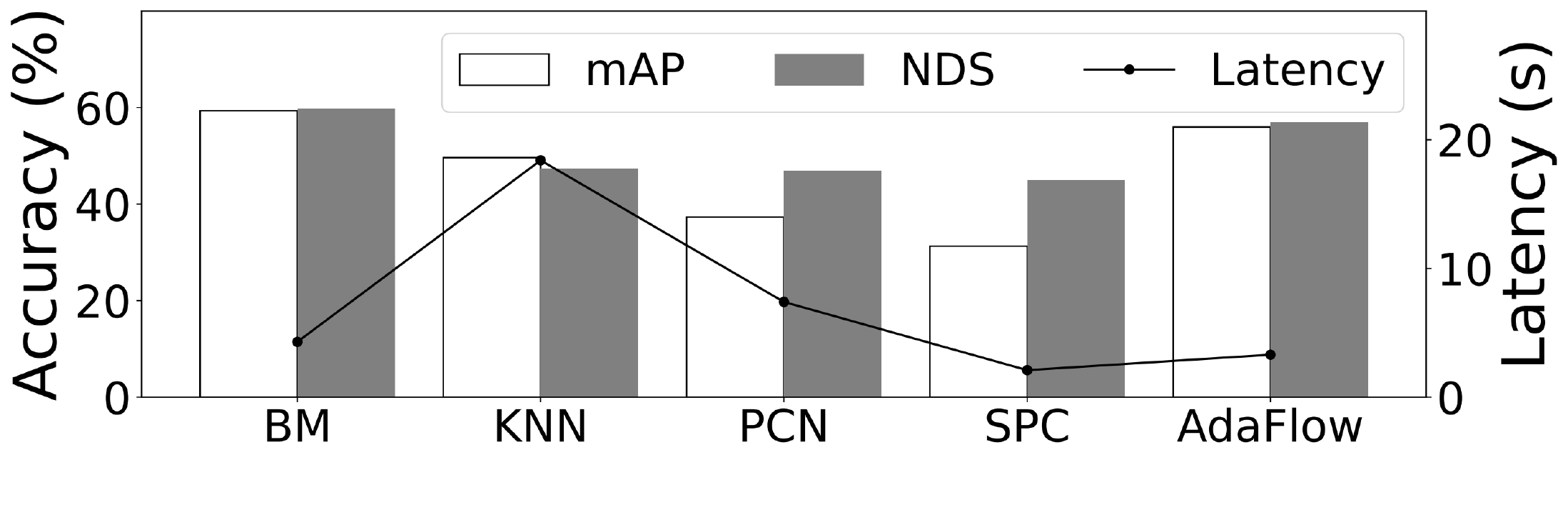}
    \label{fig:bev1}
  }
  \\
  \vspace{-3mm}
  \subfloat[BEVFusion, missing 75\%  LiDAR.]{
    \includegraphics[height=0.1\textwidth]{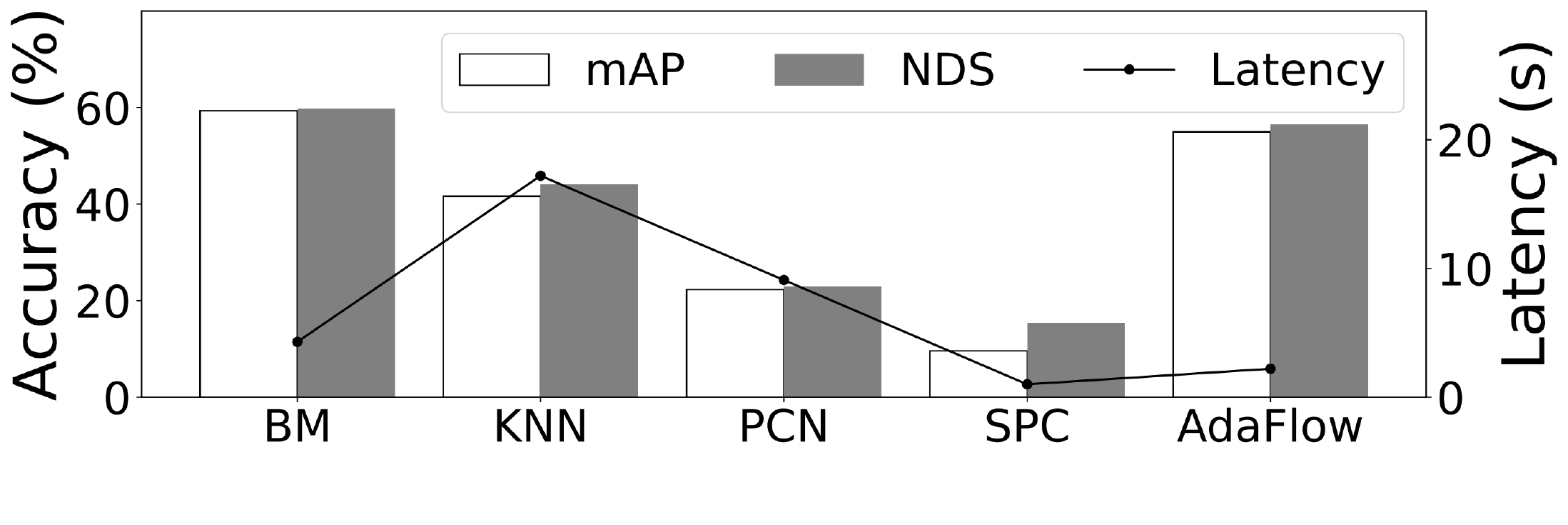}
    \vspace{-2mm}
    \label{fig:bev2}
    \vspace{-2mm}
  }
\\
\vspace{-3mm}
    \subfloat[PointPillars, missing 25\% LiDAR.]{
    \includegraphics[height=0.1\textwidth]{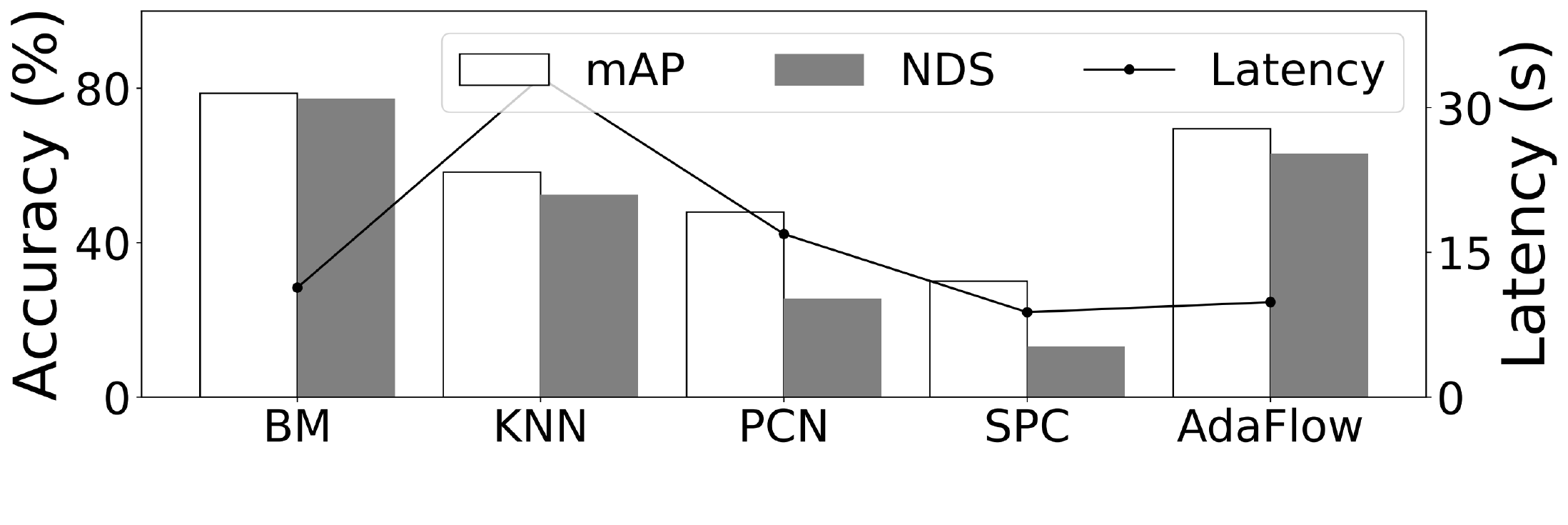}
    \label{fig:poin1}
  }
\\
\vspace{-3mm}
  \subfloat[PointPillars, missing 50\% LiDAR.]{
    \includegraphics[height=0.1\textwidth]{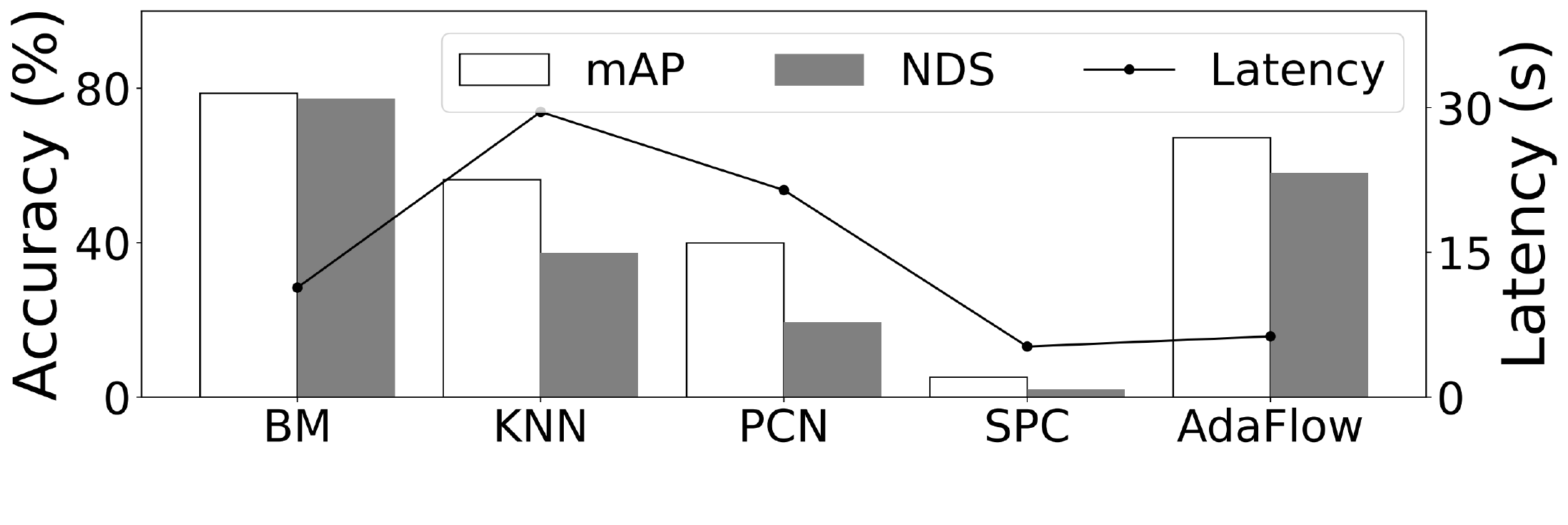}
    \label{fig:poin2}
  }
  \caption{
  Comparison of accuracy and inference latency between \sysname and baselines in two tasks, under varying missing rates of slow data (\ie lagging delay).}
  \vspace{-3mm}
  \label{fig:exp:bevpoin}
\end{figure}

\figref{fig:exp:bevpoin} shows the results.
\textit{First}, \sysname exhibits the best trade-off between inference accuracy and latency compared to the baselines.
\textit{Second}, \sysname outperforms the three asynchronous fusion methods (SPC, PCN, and KNN) under various missing rates of slow data. 
For example, as shown in \figref{fig:bev2}, with 75\% missing rate in BEV, \sysname demonstrates accuracy improvements of 45.4\%, 32.7\%, 13.3\% compared to SPC, PCN, and KNN, respectively.
\textit{Third}, with acceptable accuracy, \sysname achieves the lowest latency compared to other baselines. 
\lsc{As shown in \figref{fig:poin2}, with 50\% missing rate of slow LiDAR data in PointPillars, \sysname achieves 44.5\%, 78.7\%, 70.6\% latency reduction compared to BM, KNN, and PCN. }
Besides, \sysname improves 62\% accuracy compared to SPC with only 1s slower than SPC for 81 frames.

\textbf{Summary.} \sysname offers the optimal balance between accuracy and latency, making it a promising solution for latency-sensitive tasks that often involve asychronous data.

\subsection{\sysnameposs Data Imputation Performance}
\begin{figure}[t]
	\centering
	\includegraphics[width=0.69\linewidth]{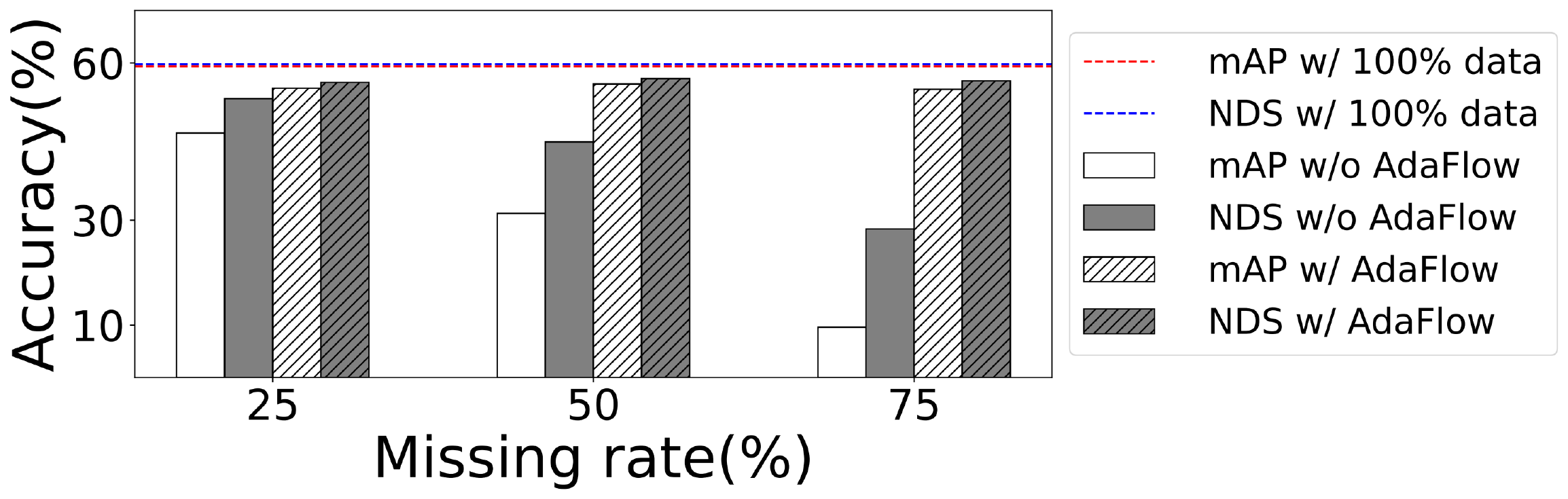}
	\caption{AdaFlow's imputation performance in BEVFusion under different missing rates of slow data.}
 \vspace{-3mm}
	\label{fig:BEV}
\end{figure}

\begin{figure}[t]
	\centering
	\includegraphics[width=0.69\linewidth]{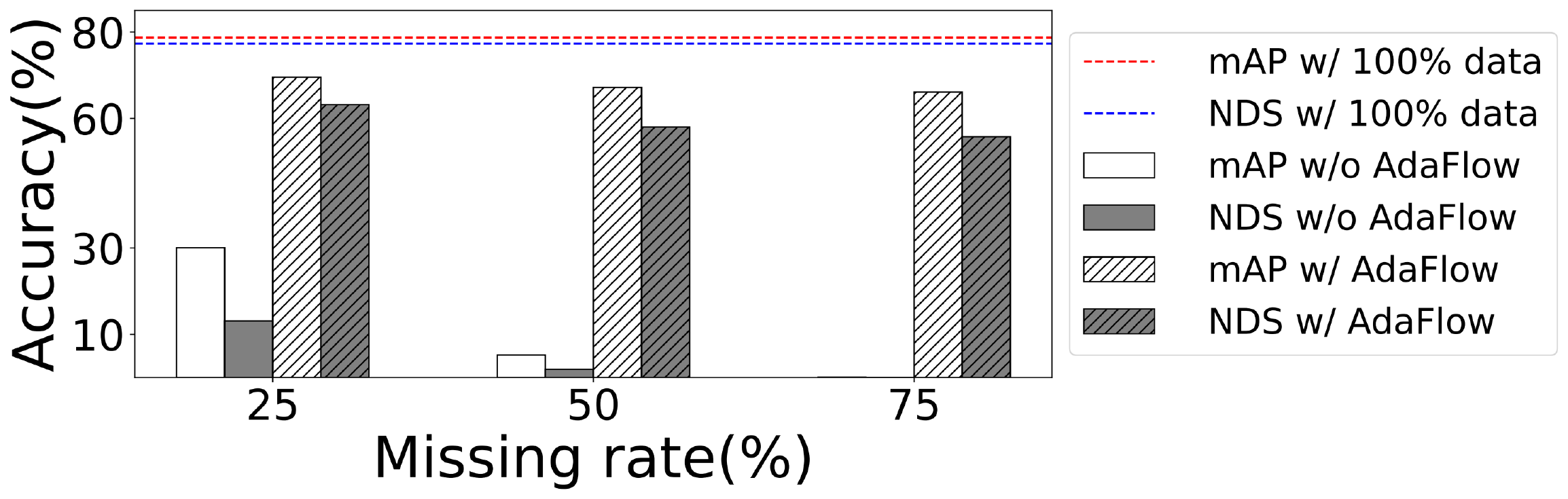}
         \vspace{-3mm}
	\caption{AdaFlow's imputation performance in PointPillars under different missing rates of slow data.}
 \vspace{-3mm}
	\label{fig:POIN}
\end{figure}

We evaluate \sysname's data imputation module on BEVFusion ($T_1$) and PointPillars ($T_2$) under different data missing rates of slow data.
\textit{First}, \sysname efficiently prevents significant accuracy decline at varying slow data missing rates.
As shown in \figref{fig:BEV}, \sysname suffer only 4.2\%, 3.4\%, and 4.4\% accuracy decline with data missing rates at 25\%, 50\%, and 75\%.
While SPC incurs up to 12.7\%, 28.1\% and 49.8\% accuracy drop, respectively.
\textit{Second}, \sysname is more necessary when suffering from \textit{extreme} slow data missing. 
As shown in \figref{fig:POIN}, \sysname improves 66\% mAP and 55.7\% NDS compared to without data imputation under data missing rates at 75\%. 
Comparatively, without \sysname's data imputation module, a 75\% data missing rate causes an 78.6\% decline in accuracy and a 77.2\% drop in NDS, making the system unusable.

\textbf{Summary.}
Under various asynchrony levels (\ie slow data missing rates), \sysname maintains system effectiveness by preventing accuracy declines. 
Particularly in extreme asychronous cases, \sysname avoids drastic accuracy drop.


\begin{figure}[t]
	\centering
	\includegraphics[width=0.69\linewidth]{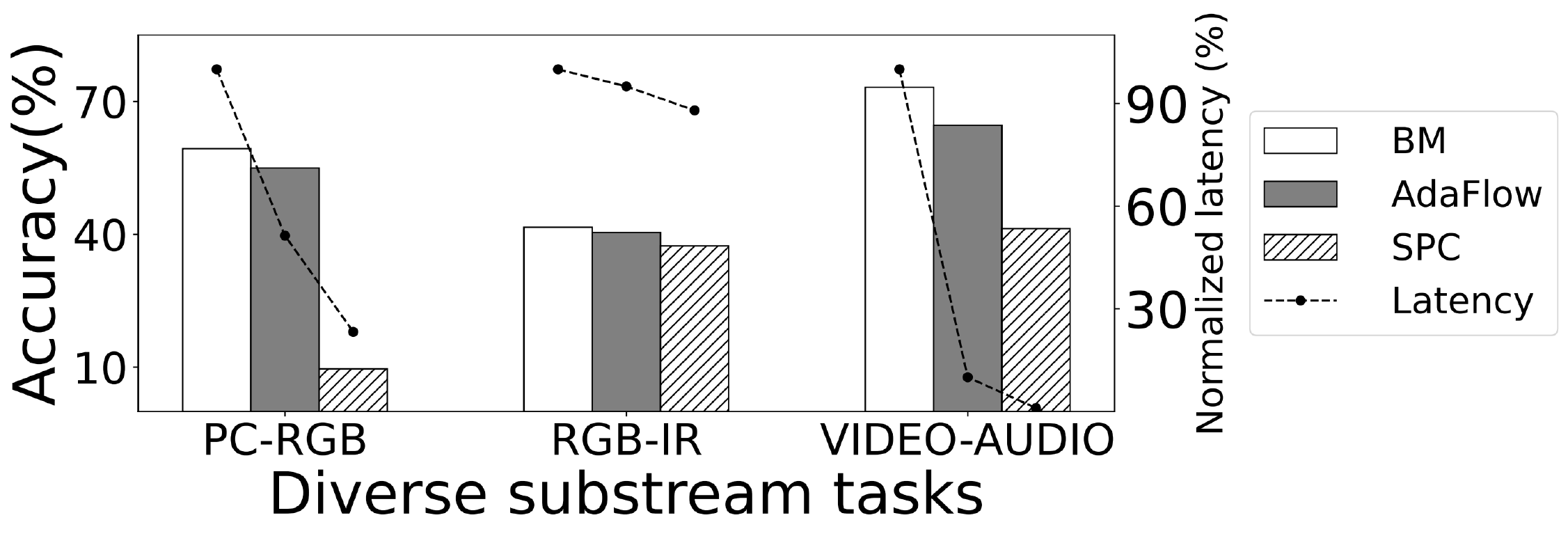}
 \vspace{-3mm}
	\caption{\lsc{Performance over diverse downstream tasks.}}
 \vspace{-3mm}
	\label{fig:gen}
\end{figure}

\subsection{Generalizing to Diverse Input Data}
We evaluate \sysname's generalization ability on three diverse tasks, \ie BEVFusion (\(T_1\)), and two LLM-based tasks ($T_{3}, T_{4}$), across different data missing rates of slow data. 
As shown in \figref{fig:gen}. \textit{First}, \sysname can adapt to various multi-modal tasks.
In detail, on \(T_1\), \(T_3\), and \(T_4\), \sysname achieve 45.4\%, 3\%, and 23.32\% accuracy improvements compared to SPC, a typical asynchronous baseline.
\lsc{\textit{Second}, \sysname performs best when the data volume ratio between fast and slow modalities is large. 
As shown in \figref{fig:acc1}.
For data ratios of 1:4, 8:9, and 1:70, \sysname reduces latency by 44.5\%, 2\%, and 90\%, respectively, compared to BM, a synchronous method with upper-bound accuracy.}
These gains are attributed to the one-fit-all ACGAN for dynamic data imputation and leverage a class attention mechanism for maximum data imputation.


\subsection{Performance of Affinity Matrix}

\subsubsection{Validity Verification}
\label{val}
\begin{figure}[t]
	\centering
	\includegraphics[width=0.7\linewidth]{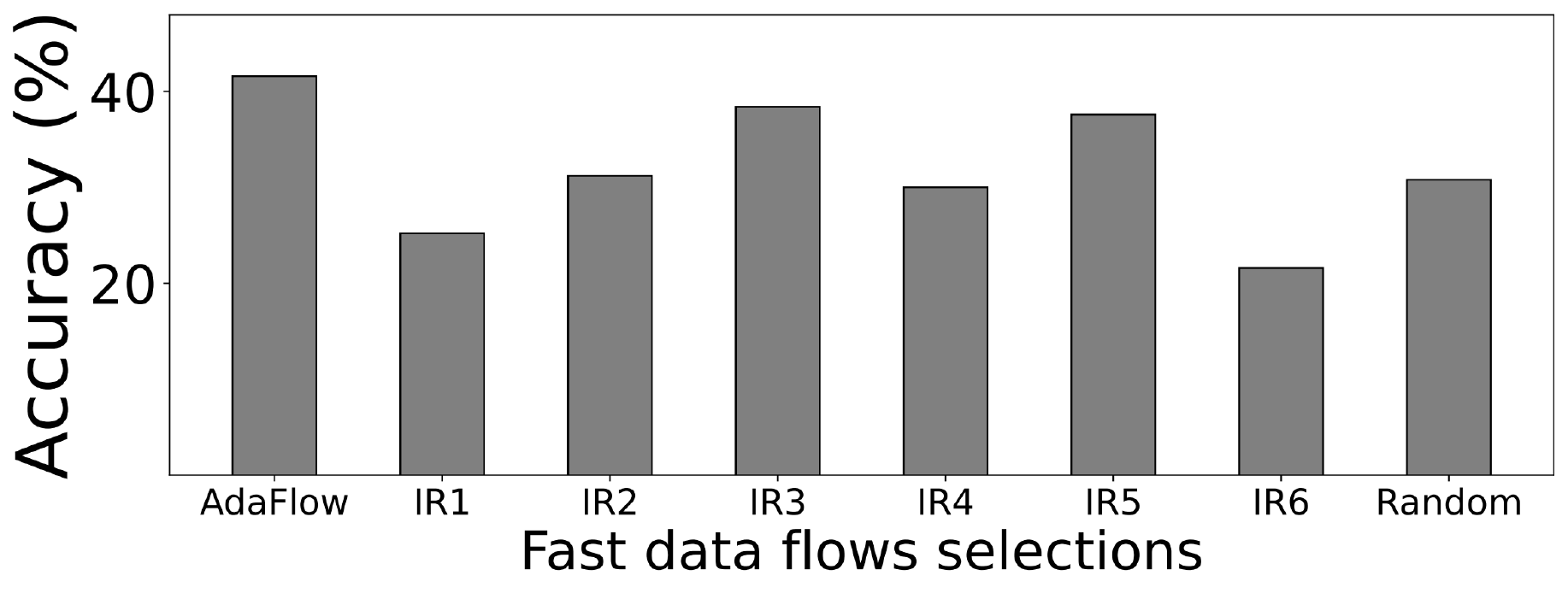}
         \vspace{-3mm}
	\caption{Comparison between \sysname and CGAN.}
         \vspace{-3mm}
	\label{fig:acc}
\end{figure}

\begin{figure}[t]
	\centering
	\includegraphics[width=0.7\linewidth]{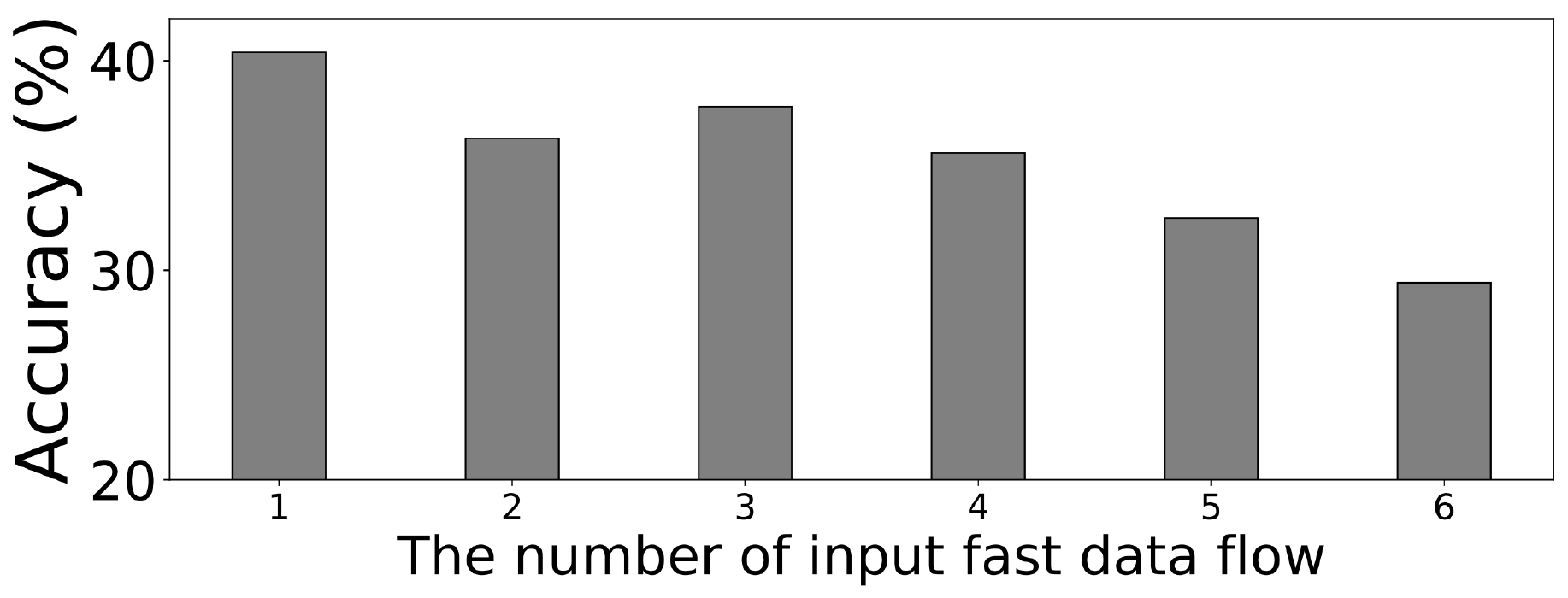}
         \vspace{-3mm}
	\caption{\lsc{Performance on diverse modality numbers.}}
         \vspace{-3mm}
	\label{fig:acc1}
\end{figure}

We test \sysname's affinity matrix on LLM task (\(T_3\)) under 10\% data missing rates of slow data. 
\textit{First}, \sysname achieves higher accuracy than CGAN's 6 fixed selections, as shown in \figref{fig:acc}. \sysname's accuracy is 40.6\% and 6 CGAN's 6 fixed selections is 36.3\%, 37.8\%, 39.6\%, 37.5\%, 39.4\% and 35.4\%, respectively. The inference derived by \sysname's generalized affinity matrix is also higher than CGAN's random selection (37.3\%). 
\lsc{\textit{Second}, 
\sysname achieves higher accuracy with just one fast modality input, as shown in \figref{fig:acc1}, while accuracy drops to 29.4\% when all modalities are used together.
This proves that the affinity matrix can always select the most suitable combination of available (fast) data for missing (slow) data imputation in diverse situations and supports the argument in \secref{graph}}.

\subsubsection{Comparison to Monte Carlo}
\begin{figure}[t]
    \centering 
     \vspace{-2mm}
    \subfloat[1st frame]{\label{fig:t1}
    \includegraphics[width=0.30\linewidth]{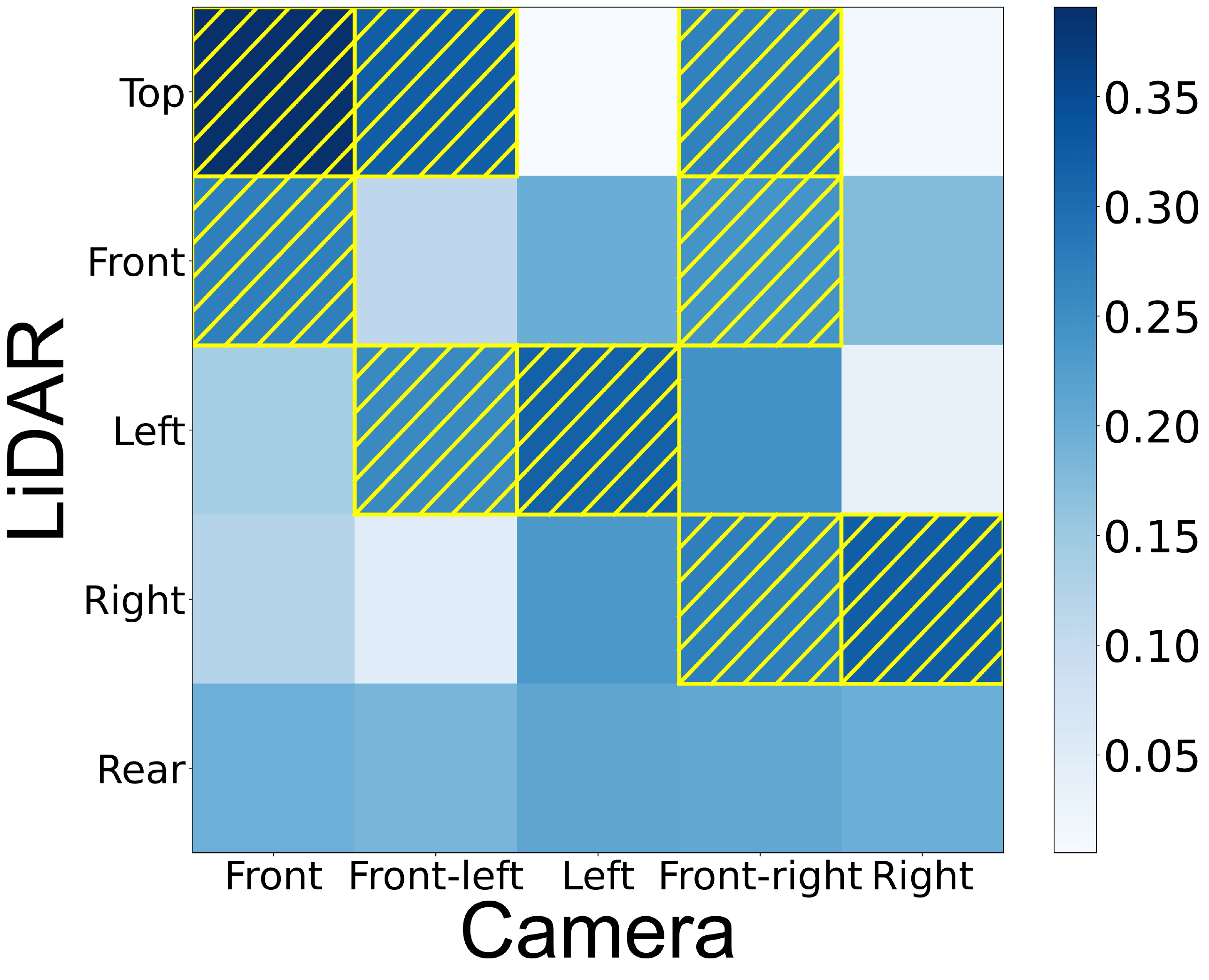}}
     \subfloat[10th frame]{\label{fig:t2}
    \includegraphics[width=0.30\linewidth]{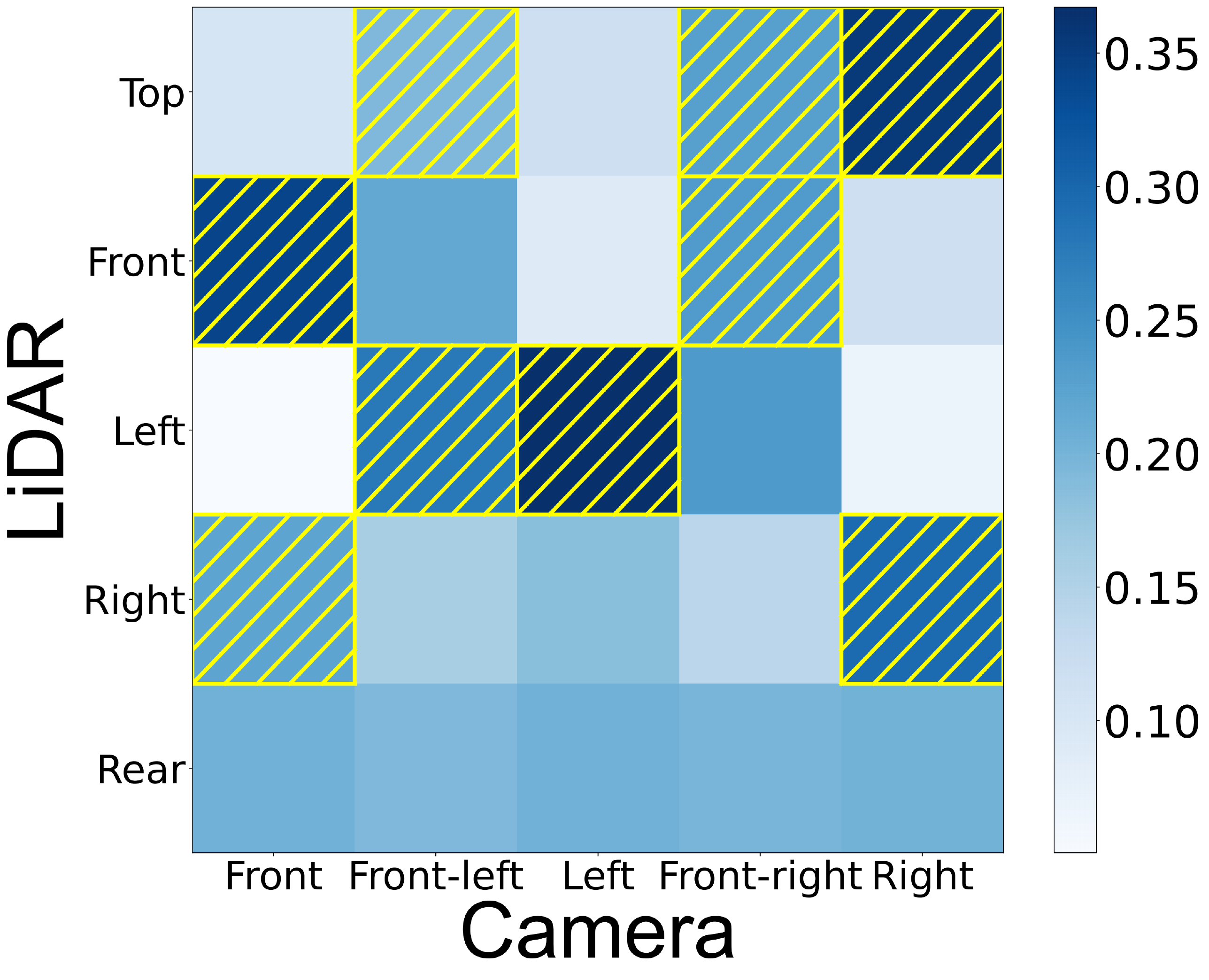}}
    \subfloat[20th frame]{\label{fig:t3}
    \includegraphics[width=0.30\linewidth]{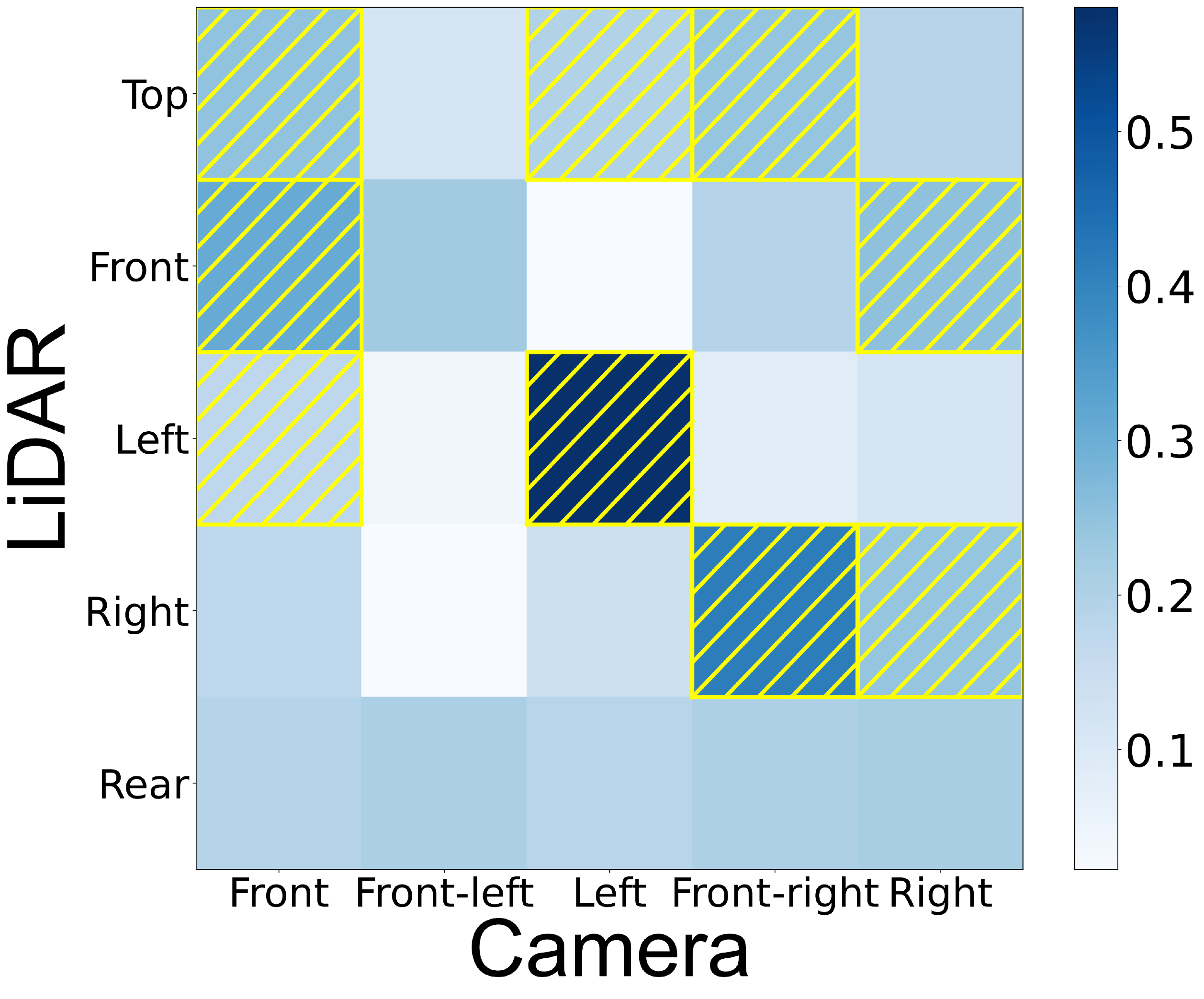}}
     \vspace{-3mm}
    \caption{Visualization of \sysnameposs affinity matrix.}
     \vspace{-3mm}
\label{fig:vm}
\end{figure}

\begin{figure}[t]
    \centering 
    \vspace{-2mm}
    \subfloat[1st frame]{\label{fig:t1g}
    \includegraphics[width=0.30\linewidth]{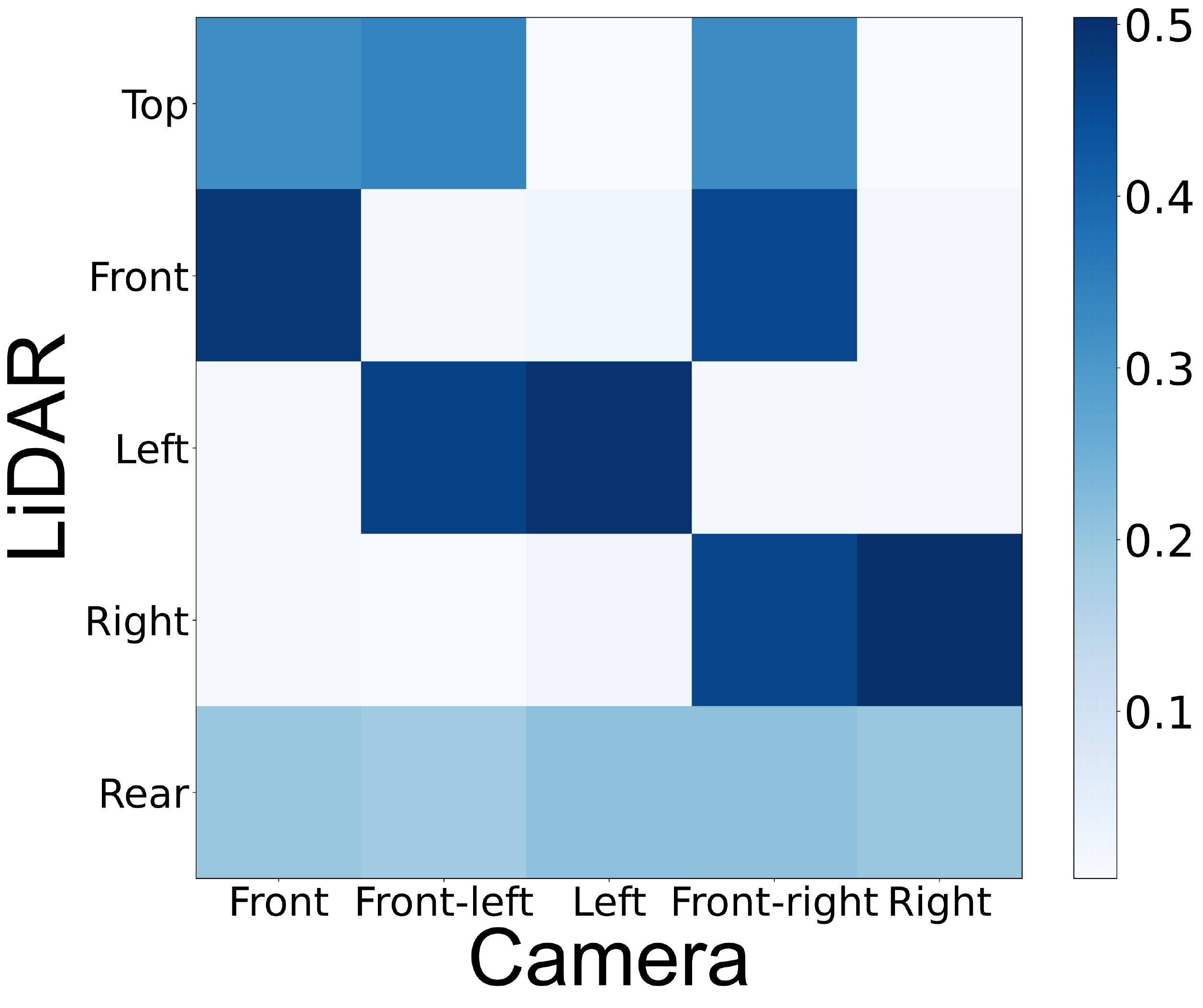}}
     \subfloat[10th frame]{\label{fig:t2g}
    \includegraphics[width=0.30\linewidth]{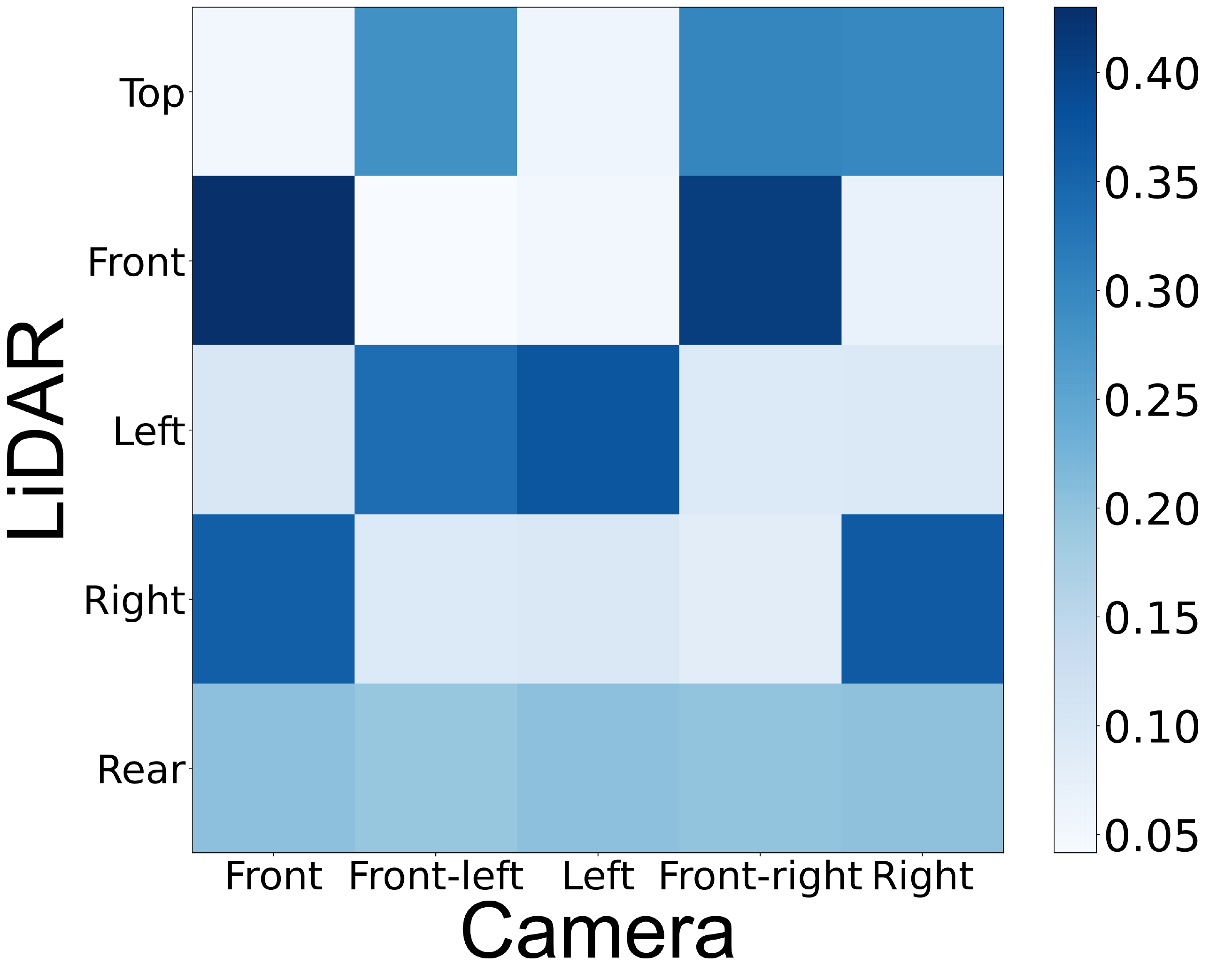}}
    \subfloat[20th frame]{\label{fig:t3g}
    \includegraphics[width=0.30\linewidth]{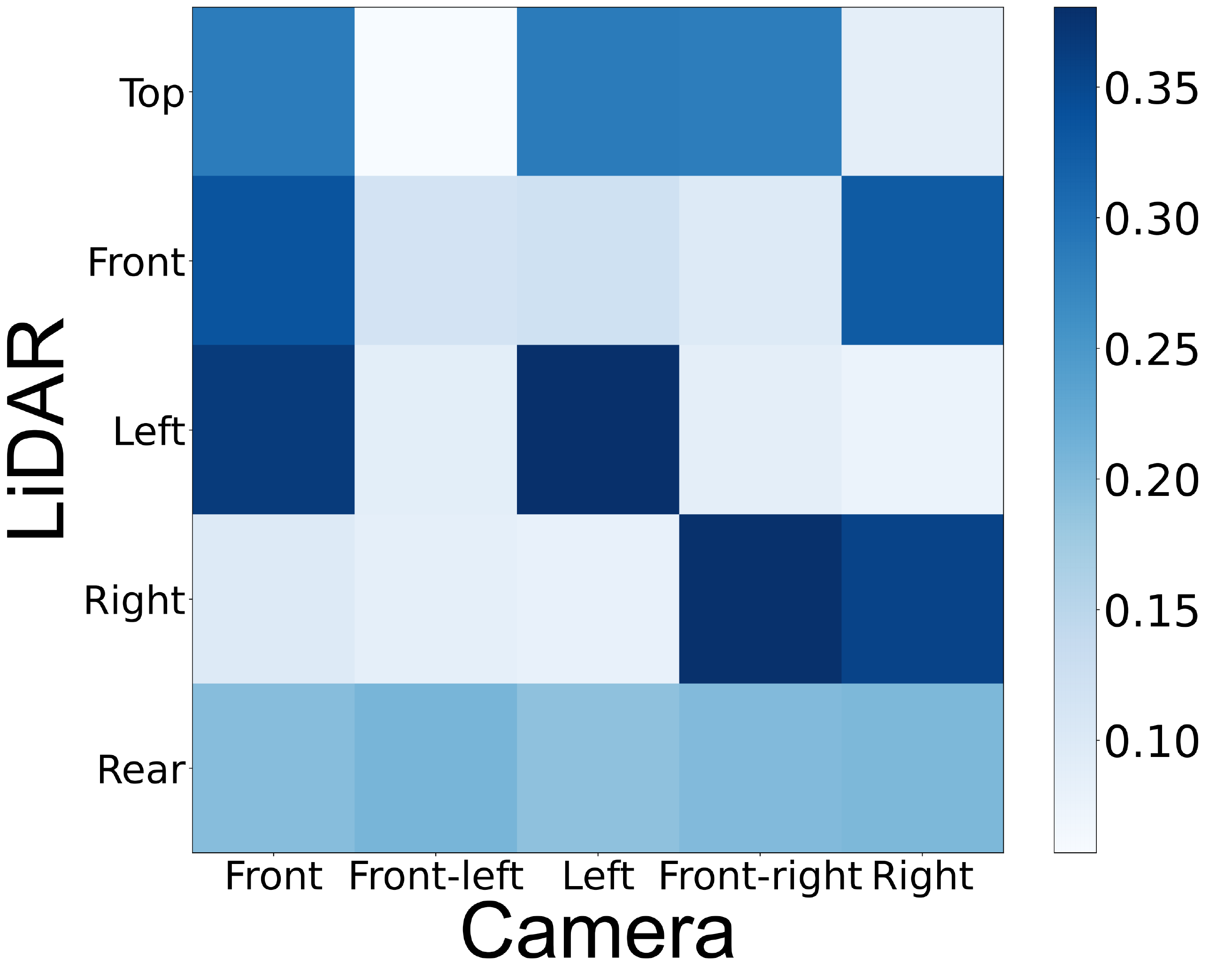}
    \label{fig:gt:c}
    }
     \vspace{-3mm}
    \caption{Visualization of expected affinity matrix.}
     \vspace{-3mm}
\label{fig:gt}
\end{figure}

We compare the affinity matrix outperformed by \sysname and Monte Carlo in \figref{fig:vm} and \figref{fig:gt}. 
For each matrix, the x-axis represents different camera views, the y-axis represents different LiDAR views. 
And elements in the matrix are the affinity value between a LiDAR view and a camera view.
In particular, we use the Monte Carlo~\cite{bib:mc:v} method here to establish relationships between modalities,  which can be regarded as the baseline (\ie ground truth). 
We chose the Monte Carlo method for its efficiency in approximating complex problems through repeated random sampling, achieving reliable solutions quickly.
\textit{First}, \sysname's affinity matrix accurately captures the relationships between modalities. 
As shown in \figref{fig:t1} and\figref{fig:t1g}, \sysname's affinity matrix provides judgments on the relationships between modalities similar to the baseline, which is quantified as 0.89 using cosine similarity.
\textit{Second}, \sysname's affinity matrix can dynamically adapt to changes in inter-modal relationships. 
As shown in \figref{fig:vm}, the same modal settings exhibited different inter-modal relationships at different times, yet \sysname consistently provided similar predictions, \ie 0.89, 0.94, and 0.91 for the first frame, 10-th frame, and 20-th frame.

\subsection{Case Study}
\begin{figure}[t]
	\centering
	\includegraphics[width=0.92\linewidth]{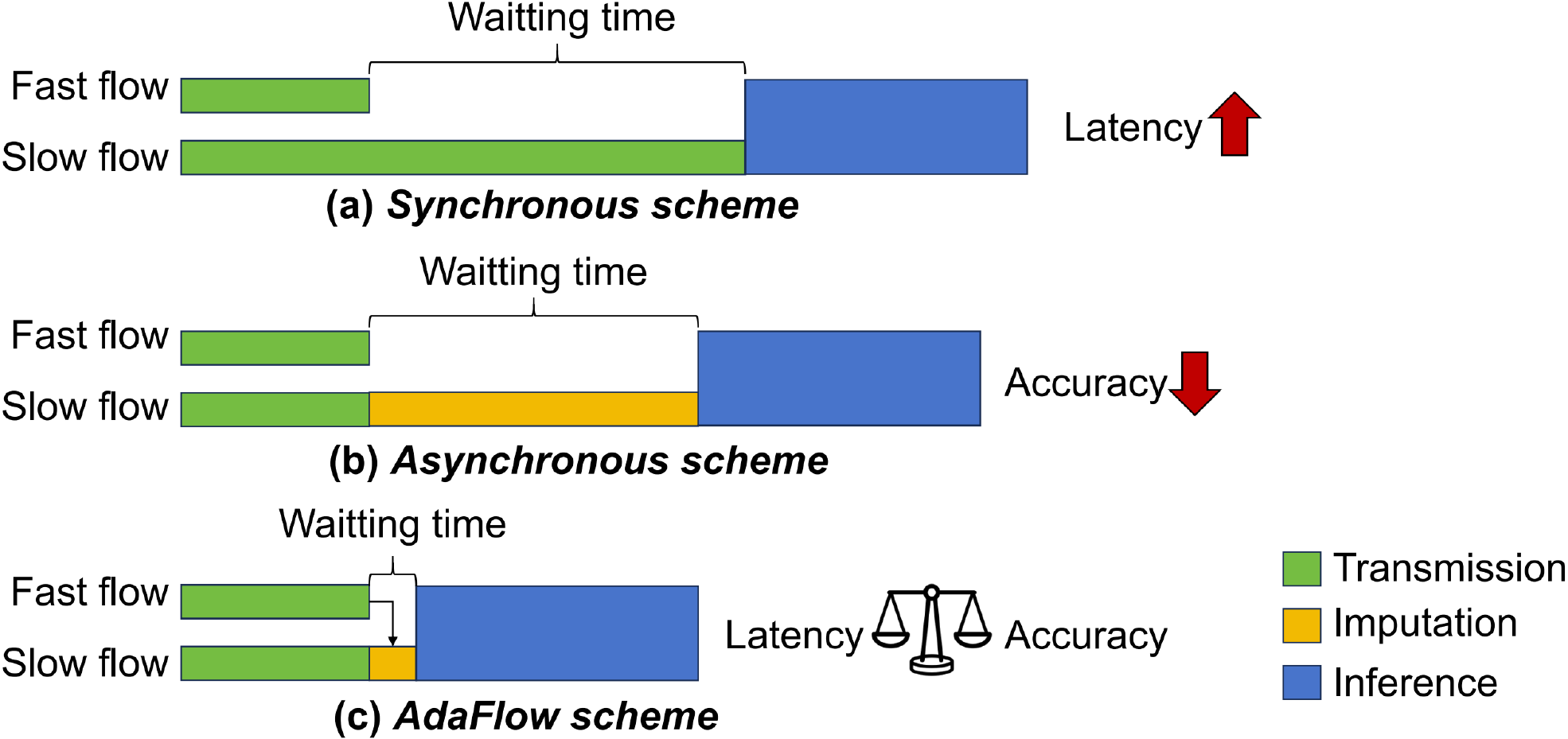}
	\caption{Paradigms of \sysname and baselines.}
	\label{fig:flow}
\end{figure}

\begin{figure}[t]
	\centering
	\includegraphics[width=0.82\linewidth]{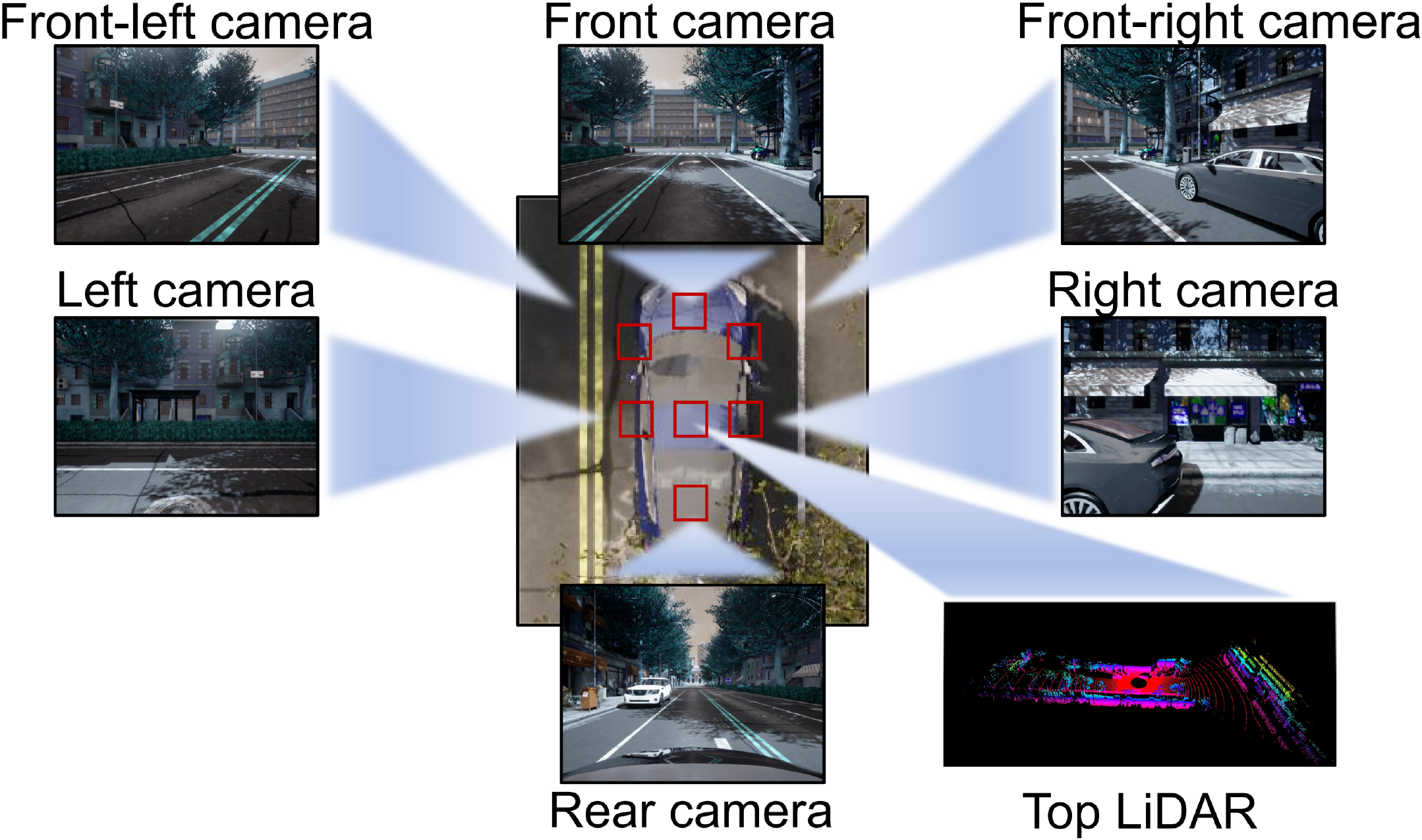}
	\caption{Illustration of seven sensors (six cameras and one LiDAR) on vehicle. }
         \vspace{-3mm}
	\label{fig:case}
\end{figure}

\figref{fig:flow} illustrates the latency breakdown in conventional end-to-end multi-modal object detection systems for autonomous driving, categorized into data preprocessing (\eg handling or imputing slow data) and executing inference. 
\sysname enhances system performance by utilizing affinity-aware data imputation for preprocessing, thereby reducing the waiting time associated with slow data and improving overall detection responsiveness.
To evaluate \sysname, we generated a 10,000-frame autonomous driving dataset using the Carla\cite{bib:carla:d} autopilot simulator. 
As demonstrated in \figref{fig:case}, the system processes inputs from six camera views (fast data) and one LiDAR view (slow data). 
We benchmark \sysname against two asynchronous methods, \ie KNN and PCN, comparing data preprocessing latency and imputation quality. 
We test data quality using Maximum Mean Discrepancy (MMD) and Chamfer Distance (CD) error metrics, comparing the imputed data with actual sensor data.

\begin{figure}[t]
    \centering 
     \vspace{-3mm}
    \subfloat[MMD]{\label{fig:c1}
    \includegraphics[width=0.30\linewidth]{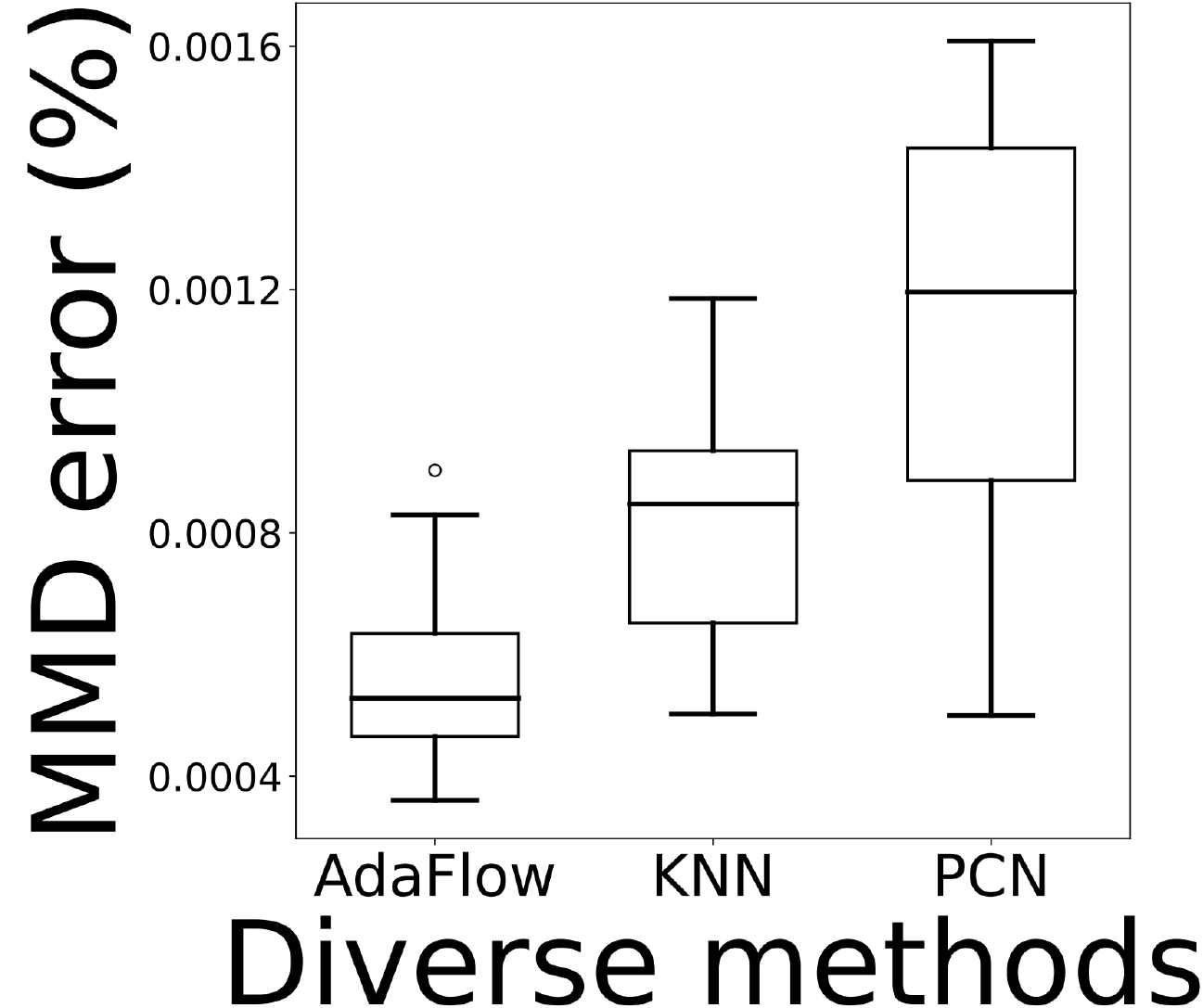}}
     \subfloat[CD]{\label{fig:c2}
     \hspace{2mm}
    \includegraphics[width=0.30\linewidth]{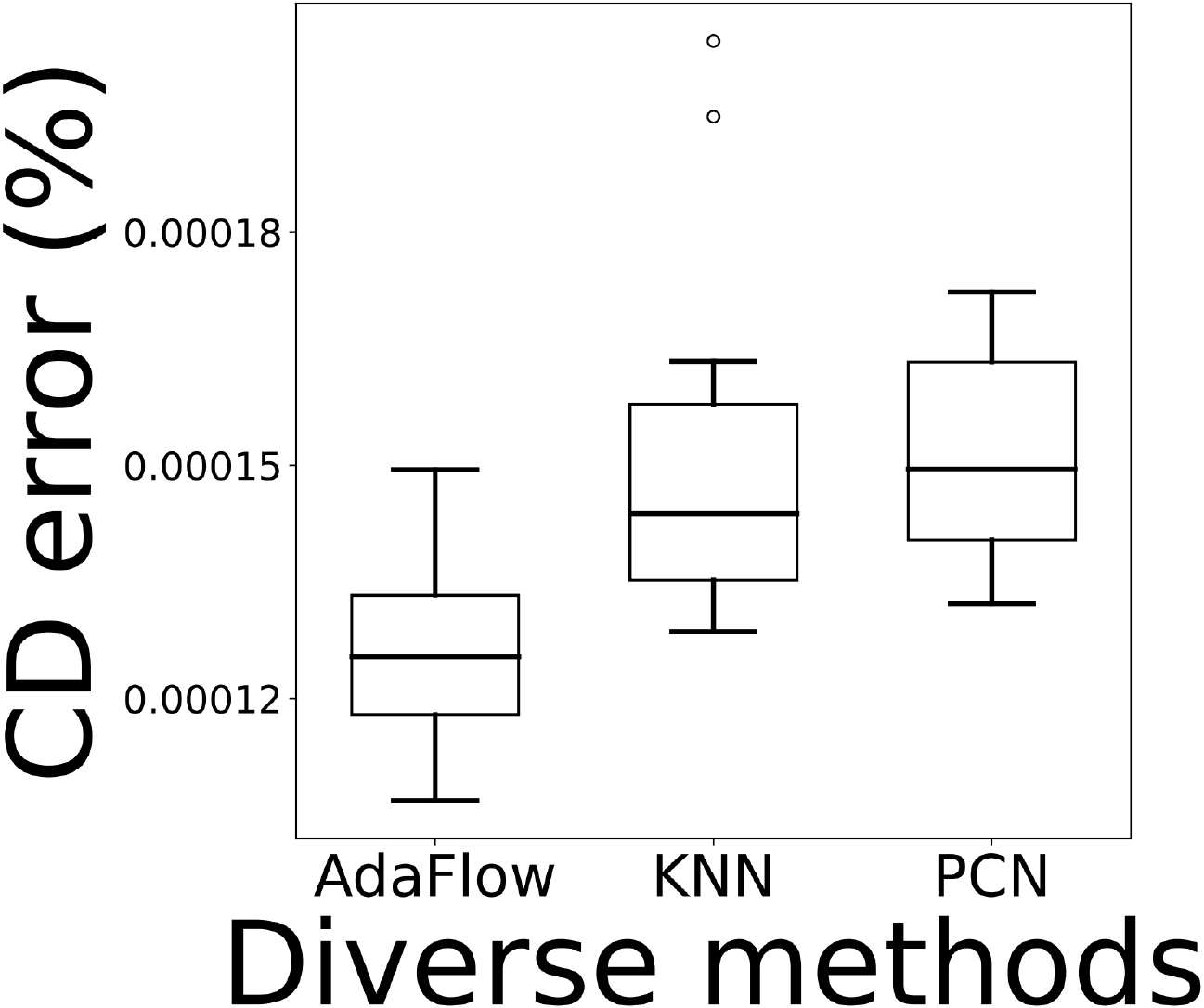}}
    \subfloat[Latency]{\label{fig:c3}
    \hspace{2mm}
    \includegraphics[width=0.30\linewidth]{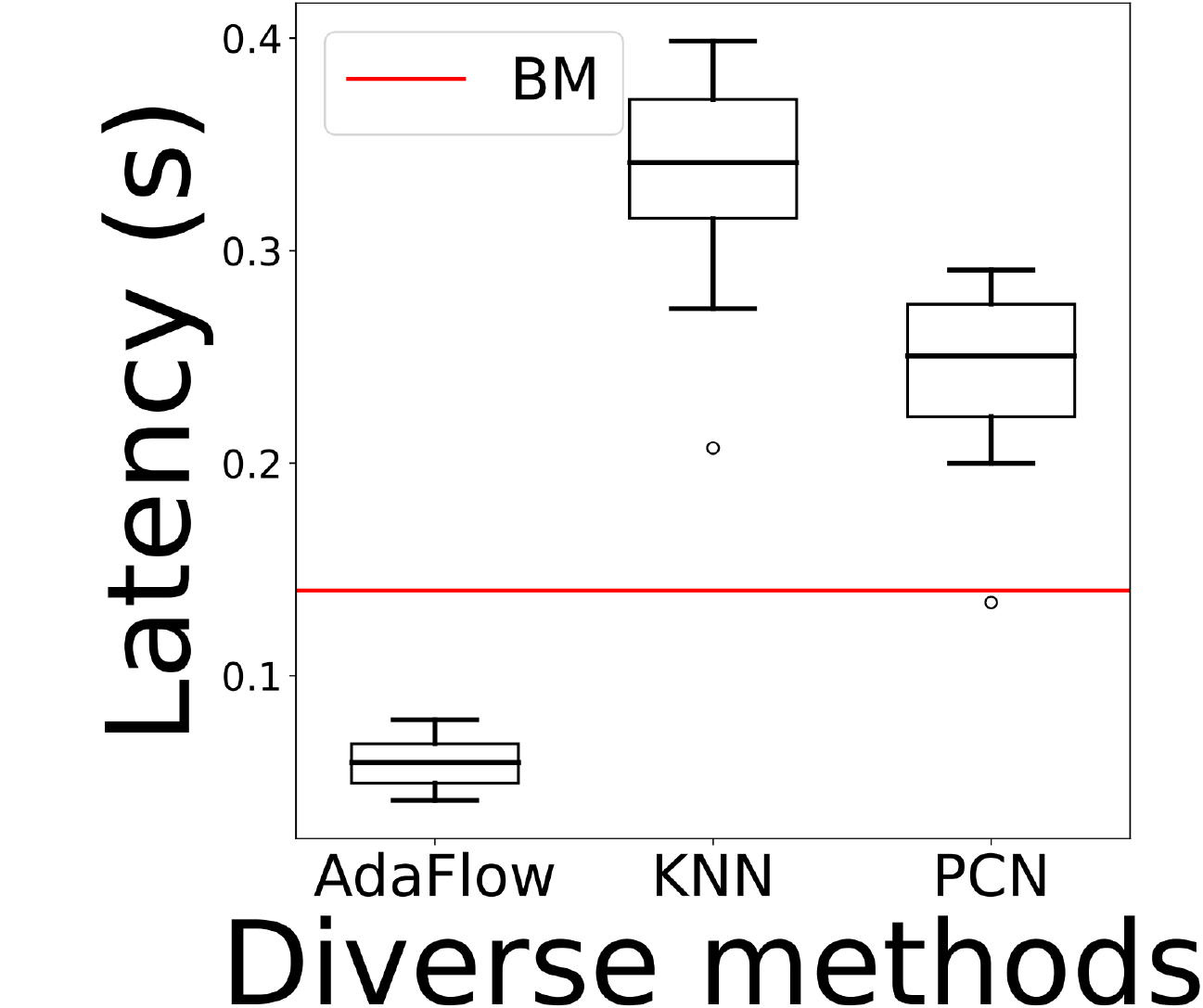}
    }
    \caption{Comparison between \sysname and asynchronous baselines (KNN, PCN) in the self-collect dataset under 66\% data missing rates of LiDAR. }
     \vspace{-3mm}
\label{fig:case1}
\end{figure}

\figref{fig:c3} shows the results.
\sysname reduces the overall latency of the preprocessing phase from 0.346s, 0.246s, and 0.141s to 0.059s, compared to KNN, PCN and BM respectively.
While when imputing data similar to those collected by real sensors (see \figref{fig:c1} and \figref{fig:c2}), \sysname achieves latency reductions of $2.2\sim 5.7\times$ than BM, PCN, and KNN. 
In practical urban settings, where vehicles commonly travel at 36 km/h, with such latency reduction, \sysname reduces the detection distance interval from 1.42m-3.46m to just 0.625m. 
This significant improvement enhances the timely detection of pedestrians and vehicles, helping accident prevention.

\begin{figure}[t]
	\centering
	\includegraphics[width=0.8\linewidth]{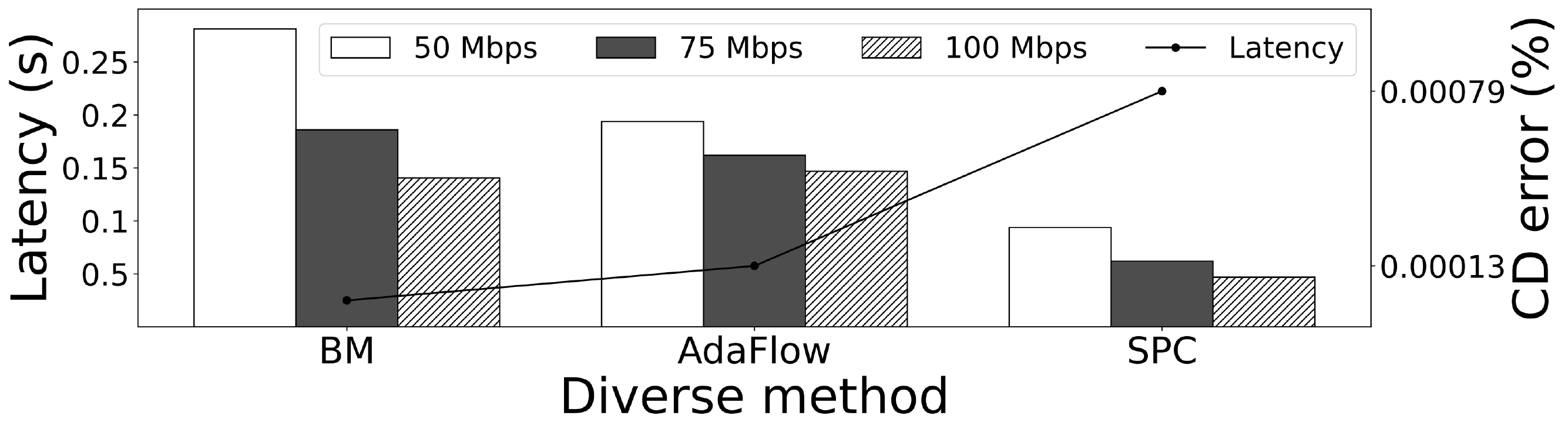}
  \vspace{-3mm}
	\caption{\lsc{Performance with diverse bandwidths.}}
  \vspace{-3mm}
	\label{fig:real}
\end{figure}

\begin{figure}[t]
 	\centering
 	\includegraphics[width=0.8\linewidth]{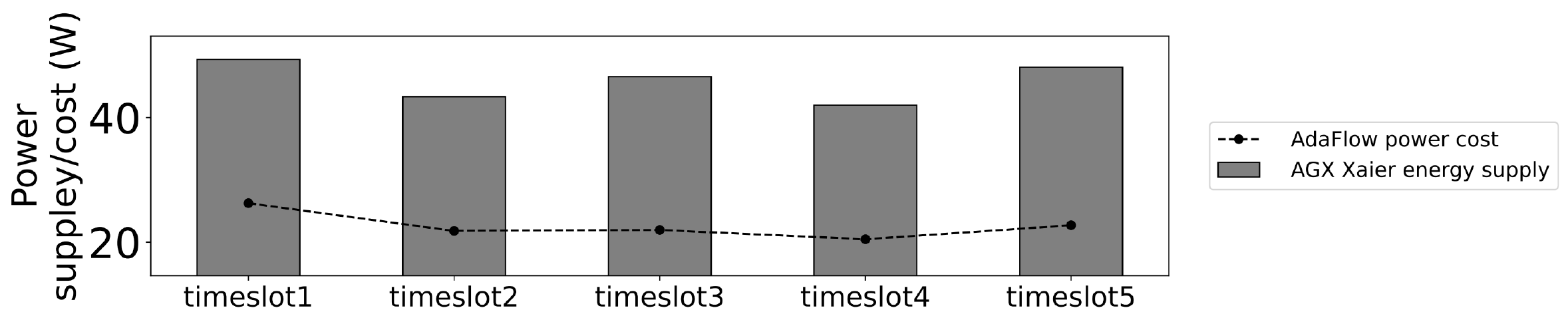}
  \vspace{-3mm}
 	\caption{\lsc{Energy cost of \sysname on AGX Xavier.}}
  \vspace{-3mm}
 	\label{fig:chips}
 \end{figure}

\lsc{
We deploy \sysname with three different bandwidths, \ie 50bps, 75Mbps, and 100Mbps on NVIDIA AGX Xaxier, similar Mercedes in-car chips. 
As shown in \figref{fig:real}, 
\sysname remains efficient even under low bandwidth. 
Compared to BM, it reduces latency by 31\% at 50Mbps while keeping CD error within acceptable limits (\ie $\leq$ 0.00015).
We also test \sysname's energy cost on the AGX Xavier. As shown in \figref{fig:chips}.
\sysname's energy cost is around 22W, which is affordable.
}

\section{Related Work}
\label{sec:related}
\textbf{Multi-modal Inference in Mobile Applications.} 
Multi-modal inference has demonstrated success across a range of mobile sensing tasks such as classification~\cite{bib:multimodal:hong,bib:multimodal:liu,3}, object detection~\cite{4,5,6,bib:spr:l,bib:evo:l,bib:caq:l}, segmentation~\cite{7,8,9}, speech recognition~\cite{10}, autonomous driving~\cite{bib:nuscenses:ca,bib:kitti:g,bib:waymo:s}, and medical image segmentation~\cite{12}. For instance, the nuScenes dataset~\cite{bib:nuscenses:ca} illustrates how integrating multi-modal data can enhance perception in autonomous driving, improving the effectiveness and user experience in driving and traffic scenarios.

\textbf{A-/Synchronous Multi-modal Inference.} 
Multi-modal inference processes data from multiple \textit{asynchronous} flows, where the slowest flow often increases waiting time. 
Existing methods are divided into \textit{synchronous blocking} and \textit{asynchronous non-blocking} approaches. 
Synchronous methods, like frame sampling~\cite{bib:fs:z}, wait for all data to arrive, optimizing latency by transmitting keyframes, but risking accuracy if slow data is low quality. 
Asynchronous methods, such as Tianxing \etal~\cite{bib:lls:L}, use predictive completion to reduce wait time but are limited by fixed input modality configurations.
\lsc{
\sysname leverages cross-modal affinity for attention-based fusion, enabling efficient asynchronous inference across diverse input modalities, regardless of number or type.
}


\textbf{Cross-modal Affinity Metrics in Multi-modal Inference.} 
The affinity metric identifies which modalities can be omitted without significantly impacting cross-modal inference accuracy. Existing approaches to quantify affinity fall into two categories: \textit{model-based} and \textit{function-based}. 
Model-based methods, like ReID~\cite{bib:reid:lu}, integrate the affinity matrix into the loss function during training, capturing affinity between specific modalities but requiring retraining for each new input. Function-based methods, such as Taskonomy~\cite{bib:taskonomy:AZ}, compute affinity from statistical data features without modifying the model. 
\sysname adopts the function-based approach for its flexibility. 
Unlike existing methods that are constrained by specific modalities, \sysname introduces a modality-agnostic affinity quantification method, enhancing cross-modal inference in dynamic, asynchronous settings.

\section{Conclusion}
To address data asynchrony and heterogeneity in distributed mobile environments, this paper introduces \sysname, which performs inference as soon as partial asynchronous data is available, improving efficiency. 
\textit{First}, \sysname models and optimizes inference using a generalized modality affinity matrix, dynamically adapting to diverse inputs for effective multi-modal fusion. It also normalizes matrix values using the analytic hierarchy process (AHP). 
\textit{Second}, \sysname employs a one-fit-all affinity attention-based conditional GAN (ACGAN) for high-accuracy, low-latency, non-blocking inference across various input modalities without retraining. 
Evaluation on real-world multi-modal tasks, \sysname significantly reduces inference latency and boosts accuracy. 
In the future, we can enhance adaptive affinity-based data imputation by leveraging the initial features of LLMs.



\begin{acks}
This work was partially supported by the National Science Fund for Distinguished Young Scholars (62025205) the National Natural Science Foundation of China (No. 62032020,
6247074224, 62102317).
\end{acks}


\bibliography{main}
\end{CJK}

\end{document}